\documentclass[11pt]{article}

\usepackage[final]{acl}
\usepackage{times}
\usepackage{latexsym}
\usepackage[T1]{fontenc}
\usepackage[utf8]{inputenc}
\usepackage{microtype}
\usepackage[hyphenate,mono]{inconsolata}
\usepackage{graphicx}
\usepackage{amsmath}
\usepackage{amssymb}
\usepackage{booktabs}
\usepackage{colortbl}
\usepackage{multirow}
\usepackage{xspace}
\usepackage{tcolorbox}
\usepackage{placeins}
\usepackage{stfloats}
\usepackage{paralist}

\definecolor{antblue}{HTML}{1677FF}

\newcommand{\shline}{\specialrule{0.6pt}{0pt}{0pt}}
\newtcolorbox{promptbox}[1]{
  colback=white,
  colframe=antblue!35,
  colbacktitle=antblue!15,
  coltitle=black,
  fonttitle=\bfseries,
  fontupper=\ttfamily,
  title={#1},
  boxrule=0.8pt,
  arc=3pt,
  boxsep=0pt,
  left=8pt,
  right=8pt,
  top=8pt,
  bottom=8pt,
  toptitle=5pt,
  bottomtitle=5pt
}

\newcommand{\method}{\mbox{\textit{CoTranslator}}\xspace}

\title{\method: Reverse to Forward Chain-of-Thought Translator for Reasoning Distillation}

\author{
 \textbf{Zinan Tang\textsuperscript{1}},
 \textbf{Zhuoshi Pan\textsuperscript{1}},
 \textbf{Kairui Chen\textsuperscript{2}},
 \textbf{Ze Yan\textsuperscript{1}},
\\
 \textbf{Shaomian Zheng\textsuperscript{2}},
 \textbf{Yukun Zhang\textsuperscript{2}},
 \textbf{Gang Wang\textsuperscript{3 $\heartsuit$}},
 \textbf{Biqing Huang\textsuperscript{1 $\heartsuit$}},
\\
 \textsuperscript{1}Tsinghua University,
 \textsuperscript{2}Ant Group,
 \textsuperscript{3}Xiaomi Corporation
\\
\small{$\heartsuit$ Corresponding Authors.}
\\[-1pt]
\small{\href{mailto:tzn26@mails.tsinghua.edu.cn}{\texttt{tzn26@mails.tsinghua.edu.cn}}; \href{mailto:hbq@tsinghua.edu.cn}{\texttt{hbq@tsinghua.edu.cn}}}
}

\begin{document}
\raggedbottom
\maketitle

\begin{abstract}
Reasoning distillation requires chain-of-thought (CoT) data with correct answers, but obtaining such data efficiently remains challenging.
Forward CoT generation (FCG) followed by rejection sampling (RS) produces verified, question-only responses, yet may require many teacher calls for difficult problems.
Reverse CoT generation (RCG) is cheaper because it conditions on the known answer, but access to this answer can lead to post-hoc rationalization and a mismatch with the student's answer-free inference setting.
We propose~\method, which learns to translate reverse CoT responses toward the distribution of verified forward responses.
Trained on a small set of paired reverse and rejection-sampled forward responses, it supports large-scale generation and verification of student training data.
Across six student models evaluated on six mathematics benchmarks, \method improves average accuracy over semantic suppression by up to $5.82$ percentage points.
For two 4B students, it retains $95.0\%$ of rejection sampling's average accuracy while using $35.9\%$ of its estimated compute.
Without retraining, the same translator transfers to a new mathematics dataset, a different teacher, and STEM problems, improving average STEM accuracy by $4.88$ and $4.31$ percentage points for 4B and 8B students, respectively.
These results show that~\method offers an efficient and transferable approach to reasoning distillation.
Dataset and Models are available at this \href{https://huggingface.co/collections/Word2Li/cotranslator}{URL}.
\end{abstract}

\begin{figure}[!htb]
\centering
\includegraphics[width=\linewidth]{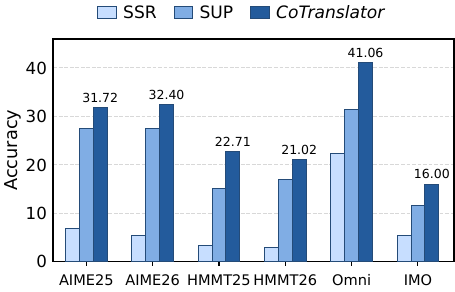}
\caption{Math benchmark accuracy of \texttt{Qwen3-4B-Base} under SSR, SUP, and~\method, using the same queries from \texttt{Nemotron-SFT-Math-v4}.}
\label{fig:main_gains}
\end{figure}

\section{Introduction}

Chain-of-thought (CoT) helps large language models (LLMs) solve complex problems by expressing intermediate reasoning steps~\citep{wei2022chain,kojima2022zero,chu2024survey}.
Reasoning distillation uses these steps to teach a smaller student model how to solve problems~\citep{shridhar2023distilling}.
The quality of this training data therefore matters, and correctness is a key requirement~\citep{yuan2023rft,lee2026answerconditioned}.
Verified question-answer pairs are relatively easy to obtain from collected mathematical data~\citep{cobbe2021verifiers,hendrycks2021math}.
However, producing and checking large scale of teacher-generated CoT responses can still be costly~\citep{shridhar2023distilling,cetin2025teachers}.
Reasoning data is therefore a bottleneck for distillation.

\begin{figure*}[!htb]
\centering
\includegraphics[width=\textwidth]{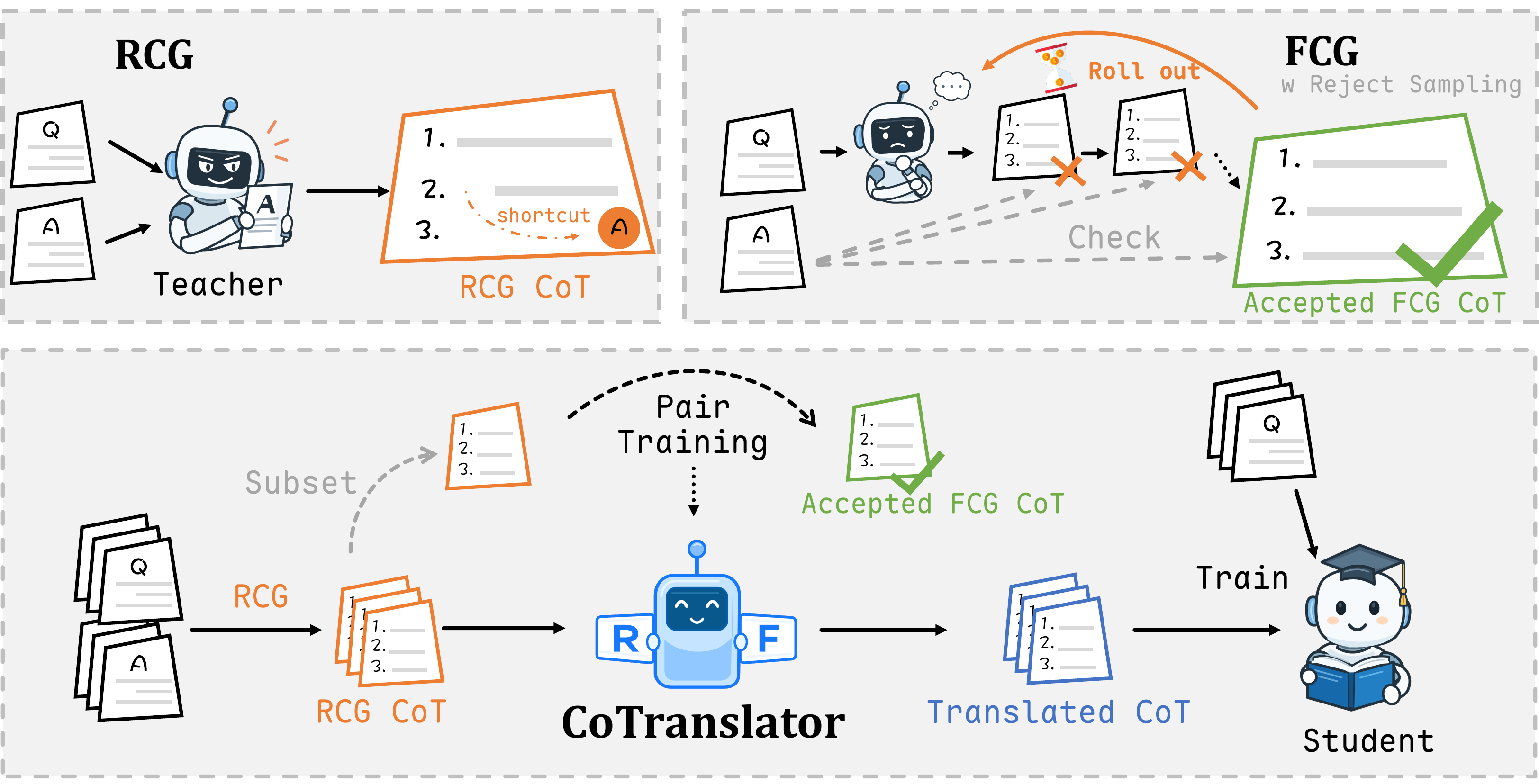}
\caption{Overview of~\method.
FCG generates reasoning from questions alone; rejection sampling keeps correct responses but may require many attempts.
RCG conditions on both the question and answer, reducing sampling cost but risking post-hoc rationalization.
Trained on paired RCG and accepted FCG responses from a small subset,~\method translates RCG responses at scale for training the student.}
\label{fig:main}
\end{figure*}

A common approach is Forward Chain-of-Thought Generation (FCG) from questions alone, followed by rejection sampling to keep responses with correct answers~\citep{zelikman2022star,yuan2023rft}.
However, rejection sampling can be costly for difficult questions.
Finding one accepted response may require many attempts, and a fixed sampling budget may leave some questions unaccepted.
Reverse Chain-of-Thought Generation (RCG) can reduce sampling cost by giving the teacher both the question and its answer~\citep{wang2025reverse,cetin2025teachers}.
However, RCG may lead to post-hoc rationalization, where the teacher justifies the known answer instead of deriving it from the question~\citep{peng2026posthoc,turpin2023unfaithful,lewislim2025dynamics}.
Correct answers and fluent explanations do not guarantee data that students can learn from.~\citep{lanham2023faithfulness,agarwal2024faithfulness,paul2024making,lee2026answerconditioned}.
One reason is train-inference mismatch: RCG responses are generated with answer access, whereas the student must reason from the question alone.
Previous methods use semantic suppression or structural guidance but still give the answer, so they do not fully resolve the mismatch~\citep{wang2025reverse,peng2026posthoc} or explicitly map RCG to the distribution of rejection-sampled FCG responses for the same question.

Inspired by consistency and recovery guarantees in machine translation~\citep{alshedivat2019consistency,zhao2020languageinvariant}, we cast the gap between reverse and forward CoT as a translation problem and propose~\method.
The overall workflow is shown in~\autoref{fig:main}.
On a small set of questions, we collect paired RCG and rejection-sampled FCG responses for the same question.
The RCG response is the source, and the accepted FCG response is the target.
We train a translator to predict the target using only the RCG source response, aiming to approximate the distribution of accepted FCG responses.
At scale, one RCG teacher call and one translator call per question produce the responses for student training.

Across six students and six math benchmarks,~\method improves average accuracy over semantic suppression by up to $5.82$ percentage points, as shown for \texttt{Qwen3-4B-Base}~\citep{yang2025qwen3} in~\autoref{fig:main_gains}.
The two 4B students retain $95.0\%$ of RS's average accuracy at $35.9\%$ of its estimated compute.
Without retraining, the translator transfers to a new math dataset, a different teacher, and STEM problems, with average STEM gains of $4.88$ and $4.31$ percentage points for 4B and 8B students.
Analyses further show greater similarity to RS responses and fewer answer-dependent shortcuts.

These results show that~\method provides an efficient and effective paradigm for reasoning distillation across models and domains.

\section{Methodology}
\label{sec:method}

We first define the source and target distributions, then describe how we construct training data for the translator and the student model.

\subsection{Source and target distributions}
\label{sec:distributions}

Let $q$ be a question, $a$ its correct answer, and $F(y)$ the final answer extracted from a response $y$.
For a teacher model $\mathcal{T}$ and a specific prompt $s$, RCG uses the answer conditioned input $(q,a)$:
\begin{equation}
  P_{\mathrm{RCG}}(x\mid q,a)=P_{\mathcal{T}}(x\mid s,q,a).
  \label{eq:rcg}
\end{equation}
FCG uses the question-only input $q$:
\begin{equation}
  P_{\mathrm{FCG}}(z\mid q)=P_{\mathcal{T}}(z\mid q).
  \label{eq:fcg}
\end{equation}
Rejection sampling keeps FCG responses like:
\begin{equation}
  P_{\mathrm{RS}}(z\mid q,a)
  =\frac{P_{\mathrm{FCG}}(z\mid q)\mathbf{1}[F(z)=a]}
  {Z_{\mathcal{T}}(q,a)}.
  \label{eq:rs}
\end{equation}
Here $Z_{\mathcal{T}}(q,a)=\Pr_{Z\sim P_{\mathrm{FCG}}(\cdot\mid q)}[F(Z)=a]$ is the probability that FCG answers correctly.

After constructing the translator $\mathcal{G}_\phi(z\mid x)$, let $\widetilde P_\phi(x,z\mid q,a)$ denote the joint distribution of an RCG source $x$ and its translation $z$ for a given $(q,a)$.
The translator's probability is
\begin{equation*}
  \mathcal{G}_\phi(z\mid x)=\widetilde P_\phi(z\mid x,q,a)
  =\frac{\widetilde P_\phi(x,z\mid q,a)}
        {P_{\mathrm{RCG}}(x\mid q,a)}.
\end{equation*}
Summing over these sources gives the output distribution, which we aim to match to $P_{\mathrm{RS}}(\cdot\mid q,a)$:
\begin{equation}
  \begin{aligned}
    &\widetilde P_\phi(z\mid q,a)\\
    &\quad=\sum_x P_{\mathrm{RCG}}(x\mid q,a)\mathcal{G}_\phi(z\mid x)\\
    &\quad=\sum_x P_{\mathrm{RCG}}(x\mid q,a)
      \frac{\widetilde P_\phi(x,z\mid q,a)}
           {P_{\mathrm{RCG}}(x\mid q,a)}\\
    &\quad=\sum_x\widetilde P_\phi(x,z\mid q,a)
      \approx P_{\mathrm{RS}}(z\mid q,a).
  \end{aligned}
  \label{eq:marginal}
\end{equation}

Detailed mathematical analysis for distribution recovery are provided in~\autoref{sec:mathematical_analysis}.

\subsection{Data construction, training and inference}

We collect small but enough training pairs to train the translator to convergence.
For each question, draw $x\sim P_{\mathrm{RCG}}(\cdot\mid q,a)$ and $z_j\sim P_{\mathrm{FCG}}(\cdot\mid q)$.
Keep the first response satisfying $F(z_j)=a$.
The resulting paired set is $\mathcal D_{\mathrm{pair}}=\{(q_i,a_i,x_i,z_i)\}_{i=1}^{m}$.

The translator is fine-tuned on these pairs, using $x_i$ as the input and the accepted response $z_i$ as the target.
With teacher forcing objective:
\begin{equation}
  \widehat{\mathcal L}(\phi)=-\frac{1}{m}\sum_{i=1}^{m}\sum_{t=1}^{|z_i|}\log \mathcal{G}_\phi(z_{i,t}\mid x_i,z_{i,<t}).
  \label{eq:loss}
\end{equation}
The $(q_i,a_i)$ is used for pairing and verification.
Consistent with the derivation in~\autoref{eq:marginal}, we condition the translator only on the RCG output $x_i$, excluding $(q_i,a_i)$ inputs also block a direct question-answer bypass and require the model to learn to translate $x_i$~\citep{bahuleyan2018variational}.

With the translator fixed, we then construct student training data.
Each question uses one teacher call $x$ and one translator call $\hat z\sim \widehat{\mathcal{G}}_\phi(\cdot\mid x)$ to form $\mathcal D_{\mathcal{S}}=\{(q_i,\hat z_i)\}_{i=1}^{n}$, where $n$ is the number of final training examples and $m<n$.

The student $\mathcal{S}_\theta$ is fine-tuned to generate the full translated dataset $\hat z$ from $q$:
\begin{equation}
  \widehat{\mathcal L}_{\mathcal{S}}(\theta)=-\frac{1}{n}\sum_{i=1}^{n}\sum_{t=1}^{|\hat z_i|}\log \mathcal{S}_\theta(\hat z_{i,t}\mid q_i,\hat z_{i,<t}).
  \label{eq:student}
\end{equation}

\section{Experiments}
\label{sec:experiments}

The experiments evaluate~\method on math benchmarks across student sizes and model families.
Transfer tests cover a new math dataset, a different teacher model, and the STEM domain.
Experimental details, including introductions to the datasets, models, benchmarks, and baselines, as well as training and evaluation hyperparameters, are provided in~\autoref{sec:experimental_details}.

\subsection{Experimental setup}

\paragraph{Data preparation.} For the main experiments, we train the translator on \texttt{35.7k} paired examples from a $1/8$ subset of the \texttt{cot} subset data in \texttt{nvidia/\allowbreak{}Nemotron-SFT-Math-v4}~\citep{du2025nemotronmath}.
The translator targets are taken directly from the original dataset and were generated by \texttt{DeepSeek-V4-Pro}~\citep{deepseekai2026deepseekv4}, rejection-sampled, and verified against the ground-truth answers.
Translator inputs are RCG responses generated by the same model with Semantic Suppression (SUP)~\citep{wang2025reverse,peng2026posthoc}, using the template in~\autoref{fig:rcg_prompt}.
We use the trained translator to translate SUP responses generated for the full \texttt{cot} subset of approximately \texttt{285.5k} records.
The translated responses are used for student training.

\begin{table*}[!htb]
\centering
\begin{tabular*}{\linewidth}{@{\extracolsep{\fill}}lcccccc>{\columncolor{antblue!7}[\tabcolsep][0pt]}c@{}}
\shline
\multirow{2}{*}{\textbf{Method}} & \textbf{\texttt{AIME25}} & \textbf{\texttt{AIME26}} & \textbf{\texttt{HMMT25}} & \textbf{\texttt{HMMT26}} & \textbf{\texttt{Omni}} & \textbf{\texttt{IMO}} & \\
 & \textcolor{gray}{\texttt{pass@64\_1}} & \textcolor{gray}{\texttt{pass@64\_1}} & \textcolor{gray}{\texttt{pass@64\_1}} & \textcolor{gray}{\texttt{pass@64\_1}} & \textcolor{gray}{\texttt{pass@1\_1}} & \textcolor{gray}{\texttt{pass@8\_1}} & \multirow{-2}{*}{\textit{\textbf{Avg.}}} \\
\shline

\rowcolor{antblue!15}\multicolumn{8}{c}{\texttt{Qwen3-4B-Base}} \\
\shline
\textcolor{gray}{RS} & \textcolor{gray}{32.55} & \textcolor{gray}{33.80} & \textcolor{gray}{19.38} & \textcolor{gray}{20.12} & \textcolor{gray}{41.55} & \textcolor{gray}{15.88} & \textcolor{gray}{27.21} \\
SSR & 6.77 & 5.42 & 3.28 & 2.89 & 22.27 & 5.44 & 7.68 \\
SUP & 27.45 & 27.45 & 15.16 & 16.90 & 31.44 & 11.62 & 21.67 \\
SUP-AUG & 23.85 & 23.44 & 13.70 & 14.73 & 30.19 & 11.03 & 19.49 \\
\shline
\textbf{\method} & 31.72 & 32.40 & 22.71 & 21.02 & 41.06 & 16.00 & \textbf{27.49} \\
\shline

\rowcolor{antblue!15}\multicolumn{8}{c}{\texttt{Qwen3-4B-Thinking-2507}} \\
\shline
\textcolor{gray}{RS} & \textcolor{gray}{50.47} & \textcolor{gray}{51.67} & \textcolor{gray}{43.85} & \textcolor{gray}{32.39} & \textcolor{gray}{52.87} & \textcolor{gray}{27.12} & \textcolor{gray}{43.06} \\
SSR & 13.65 & 11.51 & 5.63 & 8.10 & 28.64 & 8.47 & 12.67 \\
SUP & 45.62 & 46.88 & 37.24 & 29.17 & 42.52 & 19.97 & 36.90 \\
SUP-AUG & 37.40 & 40.94 & 31.56 & 24.53 & 39.41 & 17.00 & 31.81 \\
\shline
\textbf{\method} & 45.36 & 46.51 & 38.23 & 30.40 & 50.99 & 23.94 & \textbf{39.24} \\
\shline

\rowcolor{antblue!15}\multicolumn{8}{c}{\texttt{Qwen3-8B-Base}} \\
\shline
\textcolor{gray}{RS} & \textcolor{gray}{40.21} & \textcolor{gray}{45.73} & \textcolor{gray}{31.46} & \textcolor{gray}{26.56} & \textcolor{gray}{47.58} & \textcolor{gray}{22.00} & \textcolor{gray}{35.59} \\
SSR & 9.06 & 8.96 & 3.80 & 4.64 & 26.06 & 6.28 & 9.80 \\
SUP & 35.00 & 37.71 & 25.00 & 21.12 & 37.65 & 15.41 & 28.65 \\
SUP-AUG & 30.42 & 32.71 & 21.04 & 18.04 & 35.59 & 13.59 & 25.23 \\
\shline
\textbf{\method} & 34.01 & 37.76 & 27.50 & 22.77 & 43.86 & 17.22 & \textbf{30.52} \\
\shline

\rowcolor{antblue!15}\multicolumn{8}{c}{\texttt{Qwen3-8B}} \\
\shline
\textcolor{gray}{RS} & \textcolor{gray}{49.06} & \textcolor{gray}{52.60} & \textcolor{gray}{44.17} & \textcolor{gray}{32.43} & \textcolor{gray}{54.36} & \textcolor{gray}{27.06} & \textcolor{gray}{43.28} \\
SSR & 12.60 & 10.78 & 5.73 & 8.52 & 28.91 & 7.94 & 12.41 \\
SUP & 43.49 & 43.80 & 32.76 & 26.94 & 42.16 & 18.84 & 34.67 \\
SUP-AUG & 38.59 & 38.54 & 27.34 & 23.63 & 39.93 & 16.94 & 30.83 \\
\shline
\textbf{\method} & 40.73 & 43.54 & 35.26 & 27.37 & 48.15 & 21.25 & \textbf{36.05} \\
\shline

\rowcolor{antblue!15}\multicolumn{8}{c}{\texttt{Qwen2.5-Math-7B}} \\
\shline
\textcolor{gray}{RS} & \textcolor{gray}{45.00} & \textcolor{gray}{48.91} & \textcolor{gray}{36.61} & \textcolor{gray}{30.40} & \textcolor{gray}{53.14} & \textcolor{gray}{28.12} & \textcolor{gray}{40.36} \\
SSR & 8.54 & 9.11 & 2.92 & 5.40 & 25.84 & 7.41 & 9.87 \\
SUP & 35.68 & 42.14 & 28.28 & 24.91 & 41.01 & 19.25 & 31.88 \\
SUP-AUG & 32.29 & 35.99 & 22.81 & 20.60 & 41.67 & 15.78 & 28.19 \\
\shline
\textbf{\method} & 34.74 & 40.52 & 28.18 & 23.86 & 45.69 & 20.31 & \textbf{32.22} \\
\shline

\rowcolor{antblue!15}\multicolumn{8}{c}{\texttt{Qwen2.5-Math-7B-Instruct}} \\
\shline
\textcolor{gray}{RS} & \textcolor{gray}{43.54} & \textcolor{gray}{48.23} & \textcolor{gray}{35.26} & \textcolor{gray}{29.73} & \textcolor{gray}{51.99} & \textcolor{gray}{28.06} & \textcolor{gray}{39.47} \\
SSR & 8.54 & 9.11 & 3.91 & 4.88 & 26.04 & 7.91 & 10.07 \\
SUP & 36.77 & 38.96 & 26.56 & 24.81 & 42.39 & 18.06 & 31.26 \\
SUP-AUG & 31.30 & 37.40 & 21.51 & 21.59 & 38.73 & 16.06 & 27.77 \\
\shline
\textbf{\method} & 35.05 & 38.49 & 26.82 & 24.24 & 46.50 & 20.84 & \textbf{31.99} \\
\shline
\end{tabular*}
\caption{Math performance across student models.
\texttt{HMMT25} and \texttt{HMMT26} refer to \texttt{HMMT November 2025} and \texttt{HMMT February 2026}.
\texttt{Omni} and \texttt{IMO} refer to \texttt{Omni-MATH} and \texttt{IMO-AnswerBench}.
The highest average scores are bold.}
\label{tab:main_result}
\end{table*}

\paragraph{Training and evaluation.} The main translator is trained from \texttt{Qwen3-4B-Thinking-2507}.
Students are \texttt{Qwen3-\{4B-Base, 4B-Thinking-2507, 8B-Base, 8B\}} and \texttt{Qwen2.5-Math-\{7B-Instruct, 7B\}}~\citep{yang2024qwen25math}.
We use \texttt{Megatron-Swift}~\citep{zhao2024swift} for training and modified \texttt{OpenCompass}~\citep{opencompass2023} for evaluation.
For math, we use \texttt{AIME \{2025, 2026\}}~\citep{maa2025aime,maa2026aime}, \texttt{HMMT \{November 2025, February 2026\}}~\citep{hmmt2025november,hmmt2026february}, \texttt{Omni-MATH}~\citep{gao2024omnimath}, and \texttt{IMO-AnswerBench}~\citep{luong2025imobench}.
Here, \texttt{pass@n\_1} denotes \texttt{pass@1} estimated from $n$ sampled responses, averaged across questions~\citep{chen2021evaluating}.
For benchmarks with few questions, we use $n>1$ to reduce generation randomness.

\begin{table*}[!htb]
\centering
\begin{tabular*}{\linewidth}{@{\extracolsep{\fill}}lcccccc>{\columncolor{antblue!7}[\tabcolsep][0pt]}c@{}}
\shline
\multirow{2}{*}{\textbf{Method}} & \textbf{\texttt{AIME25}} & \textbf{\texttt{AIME26}} & \textbf{\texttt{HMMT25}} & \textbf{\texttt{HMMT26}} & \textbf{\texttt{Omni}} & \textbf{\texttt{IMO}} & \\
 & \textcolor{gray}{\texttt{pass@64\_1}} & \textcolor{gray}{\texttt{pass@64\_1}} & \textcolor{gray}{\texttt{pass@64\_1}} & \textcolor{gray}{\texttt{pass@64\_1}} & \textcolor{gray}{\texttt{pass@1\_1}} & \textcolor{gray}{\texttt{pass@8\_1}} & \multirow{-2}{*}{\textit{\textbf{Avg.}}} \\
\shline

\rowcolor{antblue!15}\multicolumn{8}{c}{\texttt{Qwen3-4B-Base}; \texttt{DeepSeek-V4-Pro}; \texttt{ODA-Math-460k}} \\
\shline
SUP & 26.30 & 28.18 & 18.96 & 17.14 & 36.11 & 14.72 & 23.57 \\
\textbf{\method} & 30.83 & 34.27 & 23.44 & 21.59 & 42.25 & 17.62 & 28.33 \\
\shline

\rowcolor{antblue!15}\multicolumn{8}{c}{\texttt{Qwen3-8B-Base}; \texttt{DeepSeek-V4-Pro}; \texttt{ODA-Math-460k}} \\
\shline
SUP & 33.33 & 36.77 & 25.99 & 23.67 & 42.48 & 19.44 & 30.28 \\
\textbf{\method} & 34.17 & 40.52 & 28.02 & 23.34 & 45.62 & 19.97 & 31.94 \\
\shline

\rowcolor{antblue!15}\multicolumn{8}{c}{\texttt{Qwen3-4B-Base}; \texttt{Kimi-K3}; \texttt{Nemotron-SFT-Math-v4}} \\
\shline
SUP & 5.31 & 4.64 & 4.17 & 4.12 & 20.93 & 5.47 & 7.44 \\
\textbf{\method} & 29.32 & 32.76 & 19.22 & 20.31 & 40.45 & 14.94 & 26.17 \\
\shline

\rowcolor{antblue!15}\multicolumn{8}{c}{\texttt{Qwen3-8B-Base}; \texttt{Kimi-K3}; \texttt{Nemotron-SFT-Math-v4}} \\
\shline
SUP & 11.25 & 8.96 & 5.68 & 5.54 & 24.32 & 8.22 & 10.66 \\
\textbf{\method} & 35.10 & 36.15 & 23.39 & 22.77 & 43.11 & 17.56 & 29.68 \\
\shline
\end{tabular*}
\caption{Math performance when changing the dataset or the teacher model.
The blue rows specify the student model, teacher model, and dataset for each setting.
\method transfers across datasets and teacher models.}
\label{tab:math_transfer}
\end{table*}

\paragraph{Baselines.} We compare~\method against representative baselines.
SUP instructs teacher model to conceal the response during generation~\citep{wang2025reverse}, as shown in~\autoref{fig:rcg_prompt}.
SUP-AUG strengthens SUP with an explicit threat of punishment for answer leakage~\citep{xu2025bullying}, as specified in~\autoref{fig:aug_sup_prompt}.
SSR (Structural Skeleton-guided Reasoning) first generates a reasoning skeleton that specifies the role of each step, then uses it to guide the full CoT response~\citep{peng2026posthoc}.
We also report the performance of student models trained on data obtained through rejection sampling (RS) as a theoretical reference.

\subsection{Main result}

The math results are shown in~\autoref{tab:main_result}.
\method improves average math accuracy over SUP for all student models.
The 4B students match the translator in size.
Their accuracy gains over SUP are $5.82$ for \texttt{Base}, which even outperforms its counterpart trained on RS responses in average accuracy, and $2.34$ for \texttt{Thinking-2507}.
The 8B students are larger than the 4B translator and also show gains over SUP.
The gains on the \texttt{Qwen2.5-Math} family show that~\method remains effective across student model families.
SSR was originally designed for instruction-following tasks and performs poorly on reasoning data, which does not outperform SUP.
SUP-AUG also underperforms SUP, suggesting that adding threats of punishment for answer leakage can instead hurt student performance.

\subsection{Transfer experiments}
\label{sec:generalization}

The transfer experiments test generalization to a new math dataset, a different teacher model, and the STEM domain.
For these tests, we reuse the translator from the main experiments without further training on the new data.
The student models are \texttt{Qwen3-\{4B-Base, 8B-Base\}}.

\paragraph{New dataset and teacher model.}
To test transfer to a new math dataset, we use full \texttt{460k} problems from \texttt{OpenDataArena/\allowbreak{}ODA-Math-460k}~\citep{gao2025closing,cai2025opendataarena} while keeping \texttt{DeepSeek-V4-Pro} as the teacher model.
For teacher transfer, we keep \texttt{nvidia/\allowbreak{}Nemotron-SFT-Math-v4} fixed and replace the teacher model with \texttt{Kimi-K3}~\citep{kimiteam2026kimik3}.
As shown in~\autoref{tab:math_transfer},~\method improves average math accuracy over SUP on the new dataset by $4.76$ and $1.66$ for the 4B and 8B students, respectively, matching the main-experiment trend and indicating good generalization.
With \texttt{Kimi-K3}, the gains are large at $18.73$ and $19.02$.
\texttt{Kimi-K3} tends to leak answers early in its responses, potentially causing poor performance.
But \method successfully helps recover from these bad cases, as discussed in~\autoref{sec:teacher_behavior}.

\begin{table}[!htb]
\centering
\resizebox{\linewidth}{!}{%
\begin{tabular}{@{}lccc>{\columncolor{antblue!7}[\tabcolsep][0pt]}c@{}}
\shline
\multirow{2}{*}{\textbf{Method}} & \textbf{\texttt{Olympiad}} & \textbf{\texttt{SciBench}} & \textbf{\texttt{OmniScience}} & \\
 & \textcolor{gray}{\texttt{pass@1\_1}} & \textcolor{gray}{\texttt{pass@1\_1}} & \textcolor{gray}{\texttt{pass@1\_1}} & \multirow{-2}{*}{\textit{\textbf{Avg.}}} \\
\shline

\rowcolor{antblue!15}\multicolumn{5}{c}{\texttt{Qwen3-4B-Base}} \\
\shline
SUP & 24.35 & 22.43 & 9.68 & 18.82 \\
\textbf{Ours} & 24.35 & 37.93 & 8.83 & 23.70 \\
\shline

\rowcolor{antblue!15}\multicolumn{5}{c}{\texttt{Qwen3-8B-Base}} \\
\shline
SUP & 37.68 & 24.71 & 12.33 & 24.91 \\
\textbf{Ours} & 31.30 & 43.86 & 12.50 & 29.22 \\
\shline

\end{tabular}%
}
\caption{Student performance on STEM benchmarks.
\texttt{Olympiad} and \texttt{OmniScience} refer to \texttt{OlympiadBench} and \texttt{AA-Omniscience}, respectively.}
\label{tab:ood_knowledge}
\end{table}

\paragraph{New Domain.}
The domain transfer experiment tests whether the translator trained on math data generalizes to the STEM domain.
We use \texttt{221.6k} problems from the \texttt{vendor} subset of \texttt{nvidia/\allowbreak{}Nemotron-SFT-Science-v2}~\citep{nvidia2026nemotronsciencev2}.
Evaluation covers the STEM subset of \texttt{OlympiadBench}~\citep{he2024olympiadbench}, \texttt{SciBench}~\citep{wang2024scibench} and \texttt{AA-Omniscience}~\citep{jackson2025aaomniscience}.
\method improves average STEM accuracy over SUP by $4.88$ for the 4B student and $4.31$ for the 8B student, as shown in~\autoref{tab:ood_knowledge}.
These gains indicate that the translator trained on math data generalizes to the STEM.
This experiment focuses on generalization.
Training a translator on paired STEM data could further improve performance.

\section{Analysis}
\label{sec:analysis}

In this section, we analyze response similarity to RS and the reduction of post-hoc rationalization, the effects of translator and training-data size on student performance, and computational savings relative to rejection sampling.

\begin{figure*}[!b]
  \centering
  \includegraphics[width=\textwidth]{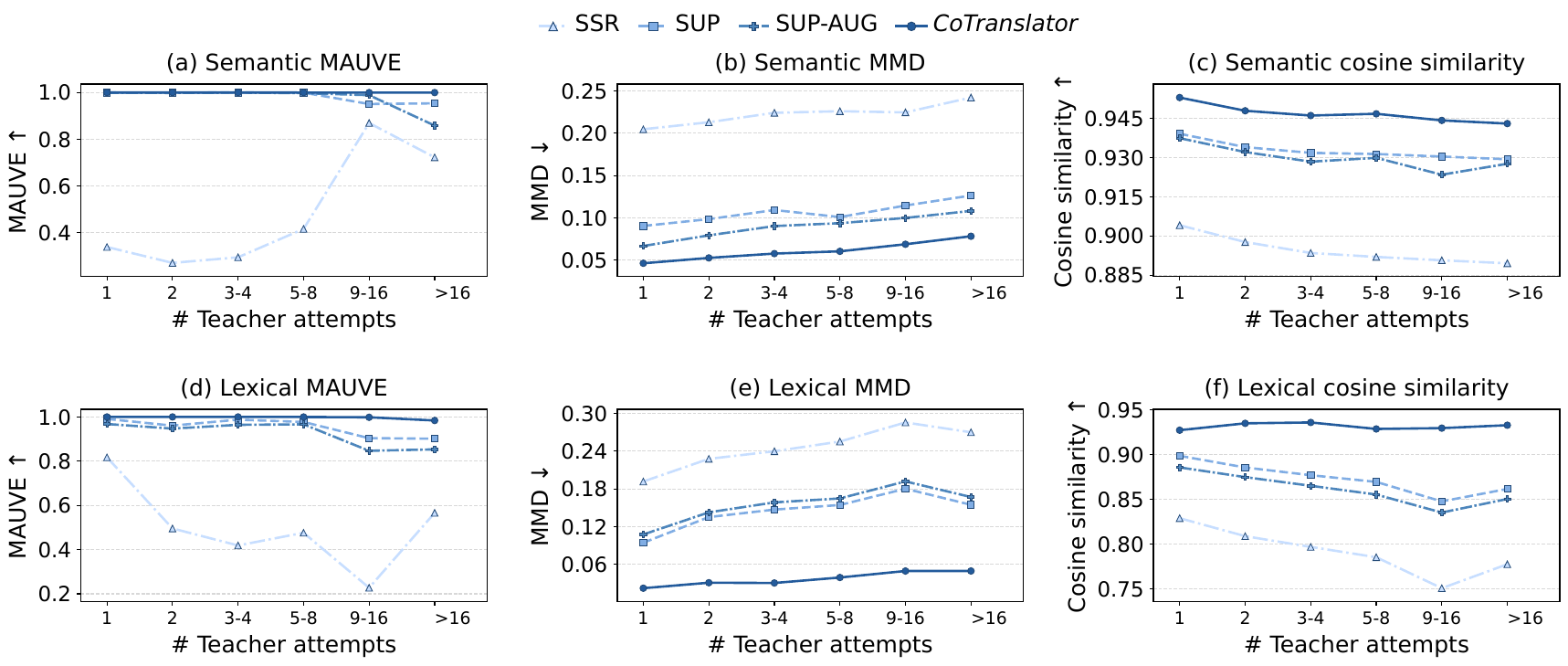}
  \caption{Similarity to RS responses by teacher attempts. Top: semantic features. Bottom: full-text lexical features. \method brings responses closer to RS in both semantic and lexical features across difficulty levels.}
  \label{fig:teacher_resemblance}
\end{figure*}

\subsection{Similarity to RS responses}
\label{sec:teacher_resemblance}

\method aims to approximate the RS response distribution and reduce repeated teacher sampling on difficult problems.
The analysis therefore compares translated and SUP responses with RS across difficulty levels.
The number of rejection sampling attempts indicates difficulty.
Rejection sampling is run on \texttt{12k} sampled problems, with up to 16 attempts per problem.
Following machine translation studies~\citep{li2023best,mitsuzawa2025mmdflagger}, we compare response distributions using MAUVE~\citep{pillutla2021mauve} and maximum mean discrepancy (MMD)~\citep{gretton2012kernel}.
Paired responses are also compared using cosine similarity.
Semantic features use \texttt{Qwen3-Embedding-4B}~\citep{zhang2025qwen3embedding} to encode all tokens of each response.
Lexical features use word and symbol counts from the full response, weighted by TF-IDF.
Measurement details are provided in~\autoref{sec:teacher_resemblance_details}.

The results in~\autoref{fig:teacher_resemblance} show that~\method has the highest MAUVE and cosine similarity and the lowest MMD in both feature types across all difficulty levels.
These results suggest that~\method brings RCG responses closer to the RS distribution, even on difficult problems.
We also measure the mean negative log-likelihood (Mean NLL) of each response under the teacher, conditioned only on the question, with results shown in~\autoref{fig:teacher_nll}.
\method has the lowest median NLL overall and across difficulty groups.

\subsection{Post-hoc rationalization}
\label{sec:teacher_behavior}

\begin{figure*}[!t]
  \centering
  \includegraphics[width=\textwidth]{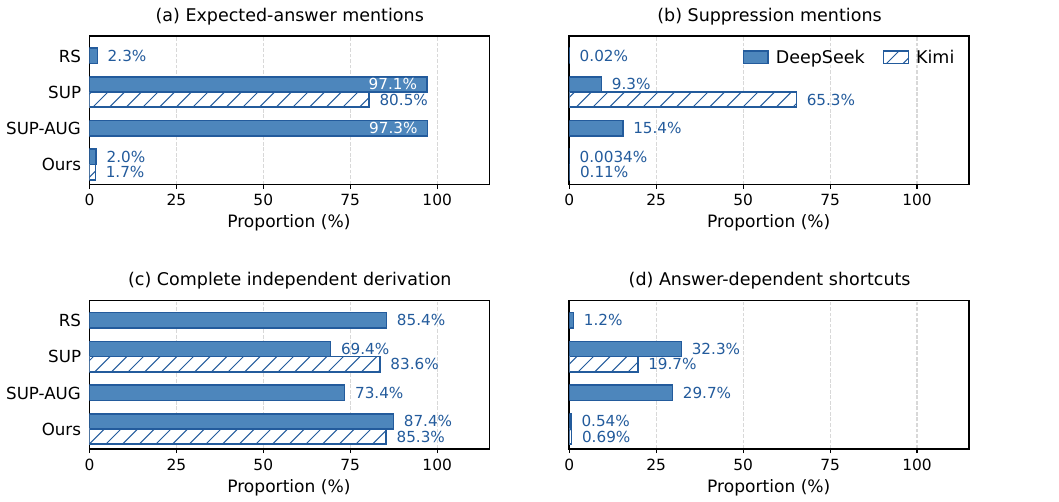}
  \caption{Post-hoc rationalization. \method reduces answer and suppression mentions and shortcuts.}
  \label{fig:teacher_behavior}
\end{figure*}

RCG can lead to post-hoc rationalization when the teacher takes shortcuts to justify the expected answer instead of deriving it from the question.
We combine rubric-based annotations, using the prompt in~\autoref{fig:teacher_behavior_rubric}, with heuristic rules to assess whether~\method reduces such shortcuts.
Specifically, \texttt{GPT-6-Luna}~\citep{openai2026gpt6systemcard} annotates acknowledgment of an externally provided expected answer (Expected-answer mentions), use of that answer to bypass required reasoning (Answer-dependent shortcuts), and self-contained derivation without using it as a premise (Complete independent derivation), while heuristic rules detect SUP prompt wording (Suppression mentions).
Following~\citet{lee2026answerconditioned}, we also measure the first occurrence of answer relative to response length (First-answer position).

\begin{figure}[!h]
  \centering
  \includegraphics[width=\linewidth]{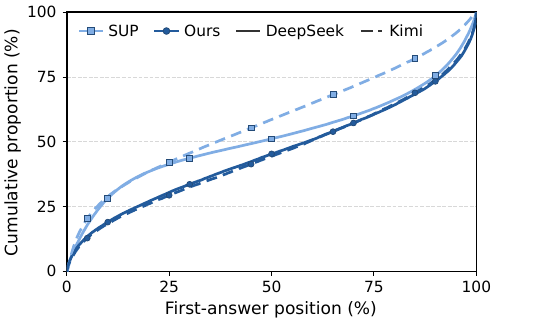}
  \caption{First-answer position in the full saved data. Translated responses mention the answer later.}
  \label{fig:teacher_first_gt}
\end{figure}

The results in~\autoref{fig:teacher_behavior} show that~\method reduces expected-answer mentions and answer-dependent shortcuts while increasing complete independent derivations for both teachers.
SUP-AUG still exhibits frequent expected-answer mentions and shortcuts despite stronger suppression instructions. The curves in~\autoref{fig:teacher_first_gt} show that the first answer appears later in the response after translation for both teachers.
Across teachers, Kimi refers to suppression instructions more and mentions the answer earlier than DeepSeek.
These tendencies may help explain the weak performance with Kimi's SUP, while the reduced reliance on answer after translation may contribute to its recovery as shown in~\autoref{tab:math_transfer}.

\subsection{Translator and training data size}
\label{sec:translator_scale}

Smaller translators and fewer paired examples can reduce the cost of distillation.
To study how translator size and training-data size affect student performance, we vary both factors to identify an efficient setting.
We use \texttt{Qwen3-\{1.7B, 4B-Thinking-2507, 8B\}} as translators.
For each model, we construct paired training data from $1/8$, $1/4$, and $1/2$ of \texttt{nvidia/\allowbreak{}Nemotron-SFT-Math-v4}, corresponding to \texttt{35.7k}, \texttt{71.4k}, and \texttt{142.8k} examples, respectively.
All training subsets are drawn from the same half of the dataset.
To ablate paired training, we also evaluate the \texttt{4B-Thinking-2507} translator without it, testing whether the model can directly refine SUP responses and remove answer-dependent shortcuts.
Each translator translates SUP responses on the other half, which is excluded from translator training.
We use the translated responses to train a \texttt{Qwen3-4B-Base} student and evaluate it on the math benchmarks.
Detailed benchmark scores are provided in~\autoref{sec:translator_scale_details}.

\begin{figure}[!h]
  \centering
  \includegraphics[width=\linewidth]{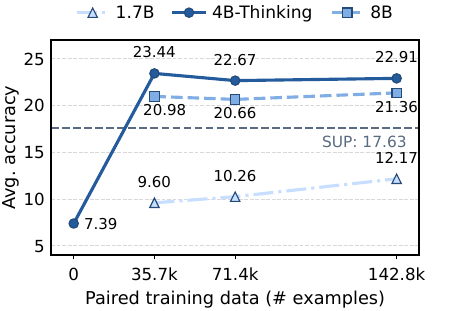}
  \caption{Math accuracy by translator size and paired training-data size. 4B and 8B show gains.}
  \label{fig:translator_scale}
\end{figure}

The results in~\autoref{fig:translator_scale} show the following:
\begin{inparaenum}[(1)]
\item The \texttt{1.7B} translator performs poorly, while the \texttt{4B} and \texttt{8B} models are large enough to learn effective trajectory translation.
\item Among the trained translators, performance varies more with model size than with training-data size.
Both larger models already perform well with \texttt{35.7k} examples.
Increasing their training data brings little improvement.
However, translation still requires paired training: using the \texttt{4B} translator without it severely reduces performance.
\item The \texttt{4B} translator consistently outperforms the \texttt{8B} translator at matched training-data sizes.
\texttt{Qwen3-4B-Thinking-2507} supports only thinking mode.
\texttt{Qwen3-8B} is a hybrid model that supports both thinking and non-thinking modes.
The focus on thinking may therefore help the \texttt{4B} model learn and generate CoT trajectories more effectively~\citep{yang2025qwen3}.
\end{inparaenum}

\subsection{Computational cost}
\label{sec:compute_resources}

\begin{figure}[!htb]
  \centering
  \includegraphics[width=\linewidth]{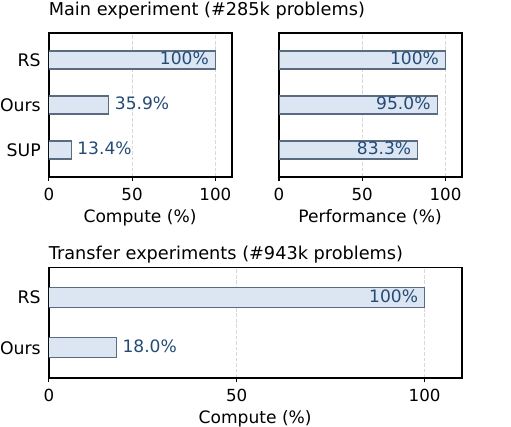}
  \caption{Computational cost and performance relative to RS. \method reduces compute by 64.1\% and 4B students achieve 95.0\% accuracy in main experiment.}
  \label{fig:compute_resources}
\end{figure}

Compute is estimated separately for the main experiment on \texttt{285k} problems from \texttt{nvidia/\allowbreak{}Nemotron-SFT-Math-v4} and the three transfer experiments on \texttt{943k} problems in total.
Following the MoE compute accounting of \citet{ludziejewski2025jointmoe}, we only use active parameter counts of 49B for \texttt{DeepSeek-V4-Pro} and 104B for \texttt{Kimi-K3}.
The main budget includes paired-data construction, translator training, teacher inference including rejection sampling, and translator inference.
Transfer experiments reuse the translator and therefore require only teacher and translator inference.
\autoref{fig:compute_resources} shows that~\method saves 64.1\% on main and 82.0\% on transfer experiments relative to rejection sampling.
With 35.9\% of RS's main-experiment compute, \method achieves 95.0\% of its mean accuracy across the two 4B students.
Details are provided in~\autoref{sec:compute_details}.

\section{Related work}
\label{sec:related_work}

\paragraph{CoT for reasoning.} CoT prompting improves reasoning through intermediate steps~\citep{wei2022chain,kojima2022zero}.
Self-consistency aggregates answers across sampled reasoning paths, while least-to-most prompting decomposes problems~\citep{wang2023selfconsistency,zhou2023leasttomost}.
Reasoning distillation trains smaller models on teacher responses through fine-tuning, task specialization, or problem decomposition~\citep{ho2023reasoningteachers,fu2023specializing,shridhar2023distilling}.
Rationale supervision reduces data requirements~\citep{hsieh2023distillingstep}, and DeepSeek-R1 demonstrates distillation into compact reasoning models~\citep{guo2025deepseekr1}.
Bootstrapping and correctness filtering further improve reasoning training data~\citep{zelikman2022star,yuan2023rft}.

\paragraph{Reverse CoT generation.} RCG reconstructs reasoning from a question and its known answer.
STaR uses answer-conditioned rationalization when forward generation fails~\citep{zelikman2022star}.
Reinforcement-Learned Teachers optimize explanations for student learning~\citep{cetin2025teachers}, while REER reconstructs reasoning for open-ended generation~\citep{wang2025reverse}.
RAVR uses a variational objective to transfer answer-conditioned reasoning to question-only generation~\citep{lin2026ravr}.
This setting raises concerns about post-hoc rationalization and CoT faithfulness~\citep{turpin2023unfaithful,lanham2023faithfulness}.
SUP instructs models to conceal the answer~\citep{wang2025reverse}, while SSR guides response generation with an answer-invariant skeleton~\citep{peng2026posthoc}.
Answer-conditioned responses can harm verifiable-reasoning distillation despite final-answer filtering~\citep{lee2026answerconditioned}.

\section{Conclusions}

We introduced~\method, which learns from a small paired set to translate RCG into reasoning that approximates rejection-sampled FCG.
It improves student performance across model sizes and families, and transfers to new datasets, teachers, and the STEM domain.
Translated responses are closer to RS and show fewer answer-dependent shortcuts.
The method saves 64.1\% of compute while achieving 95.0\% of RS mean accuracy for 4B students in the main experiment, and saves 82.0\% of compute in the transfer experiments.

\section*{Limitations}

Our experiments focus on student models up to 8B parameters and supervised fine-tuning (SFT).
The applicability of~\method to larger models and other training settings, including reinforcement learning (RL) and on-policy distillation (OPD), remains to be explored.

\bibliography{references}

@inproceedings{sutskever2014sequence,
  title = {Sequence to Sequence Learning with Neural Networks},
  author = {Ilya Sutskever and Oriol Vinyals and Quoc V. Le},
  booktitle = {Advances in Neural Information Processing Systems},
  volume = {27},
  pages = {3104--3112},
  year = {2014},
  url = {https://proceedings.neurips.cc/paper_files/paper/2014/hash/5a18e133cbf9f257297f410bb7eca942-Abstract.html}
}

@inproceedings{kim2016sequence,
  title = {Sequence-Level Knowledge Distillation},
  author = {Yoon Kim and Alexander M. Rush},
  booktitle = {Proceedings of the 2016 Conference on Empirical Methods in Natural Language Processing},
  pages = {1317--1327},
  year = {2016},
  publisher = {Association for Computational Linguistics},
  doi = {10.18653/v1/D16-1139},
  url = {https://aclanthology.org/D16-1139/}
}

@inproceedings{bahuleyan2018variational,
  title = {Variational Attention for Sequence-to-Sequence Models},
  author = {Hareesh Bahuleyan and Lili Mou and Olga Vechtomova and Pascal Poupart},
  booktitle = {Proceedings of the 27th International Conference on Computational Linguistics},
  pages = {1672--1682},
  year = {2018},
  publisher = {Association for Computational Linguistics},
  url = {https://aclanthology.org/C18-1142/}
}

@inproceedings{alshedivat2019consistency,
  title = {Consistency by Agreement in Zero-Shot Neural Machine Translation},
  author = {Maruan Al-Shedivat and Ankur Parikh},
  booktitle = {Proceedings of the 2019 Conference of the North American Chapter of the Association for Computational Linguistics: Human Language Technologies, Volume 1 (Long and Short Papers)},
  pages = {1184--1197},
  year = {2019},
  publisher = {Association for Computational Linguistics},
  doi = {10.18653/v1/N19-1121},
  url = {https://aclanthology.org/N19-1121/}
}

@inproceedings{zhao2020languageinvariant,
  title = {On Learning Language-Invariant Representations for Universal Machine Translation},
  author = {Han Zhao and Junjie Hu and Andrej Risteski},
  booktitle = {Proceedings of the 37th International Conference on Machine Learning},
  volume = {119},
  series = {Proceedings of Machine Learning Research},
  pages = {11352--11364},
  year = {2020},
  publisher = {PMLR},
  url = {https://proceedings.mlr.press/v119/zhao20b.html}
}

@article{gneiting2007proper,
  title = {Strictly Proper Scoring Rules, Prediction, and Estimation},
  author = {Tilmann Gneiting and Adrian E. Raftery},
  journal = {Journal of the American Statistical Association},
  volume = {102},
  number = {477},
  pages = {359--378},
  year = {2007},
  doi = {10.1198/016214506000001437},
  url = {https://doi.org/10.1198/016214506000001437}
}

@book{cover2006elements,
  title = {Elements of Information Theory},
  author = {Thomas M. Cover and Joy A. Thomas},
  edition = {2},
  publisher = {John Wiley \& Sons},
  year = {2006},
  doi = {10.1002/047174882X},
  url = {https://doi.org/10.1002/047174882X}
}

@inproceedings{wei2022chain,
  title = {Chain-of-Thought Prompting Elicits Reasoning in Large Language Models},
  author = {Wei, Jason and Wang, Xuezhi and Schuurmans, Dale and Bosma, Maarten and Ichter, Brian and Xia, Fei and Chi, Ed and Le, Quoc V and Zhou, Denny},
  booktitle = {Advances in Neural Information Processing Systems},
  year = {2022},
  volume = {35},
  pages = {24824--24837},
  doi = {10.52202/068431-1800},
  url = {https://proceedings.neurips.cc/paper_files/paper/2022/hash/9d5609613524ecf4f15af0f7b31abca4-Abstract-Conference.html},
}

@inproceedings{kojima2022zero,
  title = {Large Language Models are Zero-Shot Reasoners},
  author = {Kojima, Takeshi and Gu, Shixiang (Shane) and Reid, Machel and Matsuo, Yutaka and Iwasawa, Yusuke},
  booktitle = {Advances in Neural Information Processing Systems},
  year = {2022},
  volume = {35},
  pages = {22199--22213},
  doi = {10.52202/068431-1613},
  url = {https://proceedings.neurips.cc/paper_files/paper/2022/hash/8bb0d291acd4acf06ef112099c16f326-Abstract-Conference.html},
}

@inproceedings{chu2024survey,
  title = {Navigate through Enigmatic Labyrinth A Survey of Chain of Thought Reasoning: Advances, Frontiers and Future},
  author = {Zheng Chu and Jingchang Chen and Qianglong Chen and Weijiang Yu and Tao He and Haotian Wang and Weihua Peng and Ming Liu and Bing Qin and Ting Liu},
  booktitle = {Proceedings of the 62nd Annual Meeting of the Association for Computational Linguistics (Volume 1: Long Papers)},
  pages = {1173--1203},
  year = {2024},
  publisher = {Association for Computational Linguistics},
  doi = {10.18653/v1/2024.acl-long.65},
  url = {https://aclanthology.org/2024.acl-long.65/}
}

@inproceedings{shridhar2023distilling,
  title = {Distilling Reasoning Capabilities into Smaller Language Models},
  author = {Kumar Shridhar and Alessandro Stolfo and Mrinmaya Sachan},
  booktitle = {Findings of the Association for Computational Linguistics: ACL 2023},
  pages = {7059--7073},
  year = {2023},
  publisher = {Association for Computational Linguistics},
  doi = {10.18653/v1/2023.findings-acl.441},
  url = {https://aclanthology.org/2023.findings-acl.441/}
}

@misc{cobbe2021verifiers,
  title = {Training Verifiers to Solve Math Word Problems},
  author = {Karl Cobbe and Vineet Kosaraju and Mohammad Bavarian and Mark Chen and Heewoo Jun and Lukasz Kaiser and Matthias Plappert and Jerry Tworek and Jacob Hilton and Reiichiro Nakano and Christopher Hesse and John Schulman},
  year = {2021},
  eprint = {2110.14168},
  archivePrefix = {arXiv},
  primaryClass = {cs.LG},
  url = {https://arxiv.org/abs/2110.14168}
}

@inproceedings{hendrycks2021math,
  title = {Measuring Mathematical Problem Solving With the {MATH} Dataset},
  author = {Hendrycks, Dan and Burns, Collin and Kadavath, Saurav and Arora, Akul and Basart, Steven and Tang, Eric and Song, Dawn and Steinhardt, Jacob},
  booktitle = {Proceedings of the Neural Information Processing Systems Track on Datasets and Benchmarks},
  year = {2021},
  volume = {1},
  url = {https://datasets-benchmarks-proceedings.neurips.cc/paper_files/paper/2021/hash/be83ab3ecd0db773eb2dc1b0a17836a1-Abstract-round2.html},
}

@inproceedings{wang2025reverse,
  title = {Reverse-Engineered Reasoning for Open-Ended Generation},
  author = {Wang, Haozhe and Que, Haoran and Xu, Qixin and Liu, Minghao and Zhou, Wangchunshu and Feng, Jiazhan and Zhong, Wanjun and Ye, Wei and Yang, Tong and Huang, Wenhao and Zhang, Ge and Lin, Fangzhen},
  booktitle = {International Conference on Learning Representations},
  year = {2026},
  volume = {2026},
  pages = {45563--45589},
  url = {https://proceedings.iclr.cc/paper_files/paper/2026/hash/4bbeef01d9753fd5a29e9fd02d275698-Abstract-Conference.html},
}

@inproceedings{cetin2025teachers,
  title = {Reinforcement Learning Teachers of Test Time Scaling},
  author = {Cetin, Edoardo and Zhao, Tianyu and Tang, Yujin},
  booktitle = {Advances in Neural Information Processing Systems},
  year = {2025},
  volume = {38},
  pages = {107533--107567},
  publisher = {Curran Associates, Inc.},
  doi = {10.52202/085713-3586},
  url = {https://proceedings.neurips.cc/paper_files/paper/2025/hash/9a6b278218966499194491f55ccf8b75-Abstract-Conference.html},
}

@inproceedings{turpin2023unfaithful,
  title = {Language Models Don't Always Say What They Think: Unfaithful Explanations in Chain-of-Thought Prompting},
  author = {Turpin, Miles and Michael, Julian and Perez, Ethan and Bowman, Samuel},
  booktitle = {Advances in Neural Information Processing Systems},
  year = {2023},
  volume = {36},
  pages = {74952--74965},
  doi = {10.52202/075280-3275},
  url = {https://proceedings.neurips.cc/paper_files/paper/2023/hash/ed3fea9033a80fea1376299fa7863f4a-Abstract-Conference.html},
}

@inproceedings{lewislim2025dynamics,
  title = {Analysing Chain of Thought Dynamics: Active Guidance or Unfaithful Post-hoc Rationalisation?},
  author = {Samuel Lewis-Lim and Xingwei Tan and Zhixue Zhao and Nikolaos Aletras},
  booktitle = {Proceedings of the 2025 Conference on Empirical Methods in Natural Language Processing},
  pages = {29838--29853},
  year = {2025},
  publisher = {Association for Computational Linguistics},
  doi = {10.18653/v1/2025.emnlp-main.1516},
  url = {https://aclanthology.org/2025.emnlp-main.1516/}
}

@misc{lanham2023faithfulness,
  title = {Measuring Faithfulness in Chain-of-Thought Reasoning},
  author = {Tamera Lanham and Anna Chen and Ansh Radhakrishnan and Benoit Steiner and Carson Denison and Danny Hernandez and Dustin Li and Esin Durmus and Evan Hubinger and Jackson Kernion and Kamil\.{e} Luko\v{s}i\={u}t\.{e} and Karina Nguyen and Newton Cheng and Nicholas Joseph and Nicholas Schiefer and Oliver Rausch and Robin Larson and Sam McCandlish and Sandipan Kundu and Saurav Kadavath and Shannon Yang and Thomas Henighan and Timothy Maxwell and Timothy Telleen-Lawton and Tristan Hume and Zac Hatfield-Dodds and Jared Kaplan and Jan Brauner and Samuel R. Bowman and Ethan Perez},
  year = {2023},
  eprint = {2307.13702},
  archivePrefix = {arXiv},
  primaryClass = {cs.AI},
  url = {https://arxiv.org/abs/2307.13702}
}

@misc{agarwal2024faithfulness,
  title = {Faithfulness vs. Plausibility: On the (Un)Reliability of Explanations from Large Language Models},
  author = {Chirag Agarwal and Sree Harsha Tanneru and Himabindu Lakkaraju},
  year = {2024},
  eprint = {2402.04614},
  archivePrefix = {arXiv},
  primaryClass = {cs.CL},
  url = {https://arxiv.org/abs/2402.04614}
}

@inproceedings{paul2024making,
  title = {Making Reasoning Matter: Measuring and Improving Faithfulness of Chain-of-Thought Reasoning},
  author = {Debjit Paul and Robert West and Antoine Bosselut and Boi Faltings},
  booktitle = {Findings of the Association for Computational Linguistics: EMNLP 2024},
  pages = {15012--15032},
  year = {2024},
  publisher = {Association for Computational Linguistics},
  doi = {10.18653/v1/2024.findings-emnlp.882},
  url = {https://aclanthology.org/2024.findings-emnlp.882/}
}

@misc{peng2026posthoc,
  title = {Measuring and Mitigating Post-hoc Rationalization in Reverse Chain-of-Thought Generation},
  author = {Guangyue Peng and Zongchao Chen and Wen Luo and Yuntao Wen and Wei Li and Ruixiang Feng and Ran Le and Chen Yang and Zhenwei An and Yang Song and Tao Zhang and Houfeng Wang},
  year = {2026},
  eprint = {2602.14469},
  archivePrefix = {arXiv},
  primaryClass = {cs.CL},
  url = {https://arxiv.org/abs/2602.14469}
}

@misc{xu2025bullying,
  title = {Bullying the Machine: How Personas Increase {LLM} Vulnerability},
  author = {Ziwei Xu and Udit Sanghi and Mohan Kankanhalli},
  year = {2025},
  eprint = {2505.12692},
  archivePrefix = {arXiv},
  primaryClass = {cs.AI},
  url = {https://arxiv.org/abs/2505.12692}
}

@misc{lee2026answerconditioned,
  title = {Answer-Conditioned Chains of Thought Degrade Verifiable-Reasoning Distillation in Large Language Models},
  author = {Jungseob Lee and Seungyoon Lee and Suhyune Son and Dongyub Jude Lee and Sungbin Han and Sugyeong Eo and Heuiseok Lim},
  year = {2026},
  eprint = {2607.14552},
  archivePrefix = {arXiv},
  primaryClass = {cs.CL},
  url = {https://arxiv.org/abs/2607.14552}
}

@inproceedings{zelikman2022star,
  title = {{STaR}: Bootstrapping Reasoning With Reasoning},
  author = {Eric Zelikman and Yuhuai Wu and Jesse Mu and Noah D. Goodman},
  booktitle = {Advances in Neural Information Processing Systems},
  volume = {35},
  year = {2022},
  doi = {10.52202/068431-1126},
  url = {https://proceedings.neurips.cc/paper_files/paper/2022/hash/639a9a172c044fbb64175b5fad42e9a5-Abstract-Conference.html}
}

@misc{yuan2023rft,
  title = {Scaling Relationship on Learning Mathematical Reasoning with Large Language Models},
  author = {Zheng Yuan and Hongyi Yuan and Chengpeng Li and Guanting Dong and Keming Lu and Chuanqi Tan and Chang Zhou and Jingren Zhou},
  year = {2023},
  eprint = {2308.01825},
  archivePrefix = {arXiv},
  primaryClass = {cs.CL},
  url = {https://arxiv.org/abs/2308.01825}
}

@inproceedings{lin2026ravr,
  title = {{RAVR}: Reference-Answer-guided Variational Reasoning for Large Language Models},
  author = {Tianqianjin Lin and Xi Zhao and Xingyao Zhang and Rujiao Long and Yi Xu and Zhuoren Jiang and Wenbo Su and Bo Zheng},
  booktitle = {Findings of the Association for Computational Linguistics: ACL 2026},
  pages = {18757--18775},
  year = {2026},
  publisher = {Association for Computational Linguistics},
  doi = {10.18653/v1/2026.findings-acl.936},
  url = {https://aclanthology.org/2026.findings-acl.936/}
}

@inproceedings{he2024olympiadbench,
  title = {{OlympiadBench}: A Challenging Benchmark for Promoting {AGI} with Olympiad-Level Bilingual Multimodal Scientific Problems},
  author = {Chaoqun He and Renjie Luo and Yuzhuo Bai and Shengding Hu and Zhen Thai and Junhao Shen and Jinyi Hu and Xu Han and Yujie Huang and Yuxiang Zhang and Jie Liu and Lei Qi and Zhiyuan Liu and Maosong Sun},
  booktitle = {Proceedings of the 62nd Annual Meeting of the Association for Computational Linguistics (Volume 1: Long Papers)},
  pages = {3828--3850},
  year = {2024},
  publisher = {Association for Computational Linguistics},
  doi = {10.18653/v1/2024.acl-long.211},
  url = {https://aclanthology.org/2024.acl-long.211/}
}

@inproceedings{wang2024scibench,
  title = {{SciBench}: Evaluating College-Level Scientific Problem-Solving Abilities of Large Language Models},
  author = {Xiaoxuan Wang and Ziniu Hu and Pan Lu and Yanqiao Zhu and Jieyu Zhang and Satyen Subramaniam and Arjun R Loomba and Shichang Zhang and Yizhou Sun and Wei Wang},
  booktitle = {Proceedings of the 41st International Conference on Machine Learning},
  year = {2024},
  volume = {235},
  pages = {50622--50649},
  publisher = {PMLR},
  url = {https://proceedings.mlr.press/v235/wang24z.html},
}

@inproceedings{gao2024omnimath,
  title = {{Omni-MATH}: A Universal Olympiad Level Mathematic Benchmark For Large Language Models},
  author = {Gao, Bofei and Song, Feifan and Yang, Zhe and Cai, Zefan and Miao, Yibo and Dong, Qingxiu and Li, Lei and Ma, Chenghao and Chen, Liang and Xu, Runxin and Tang, Zhengyang and Wang, Benyou and Zan, Daoguang and Quan, Shanghaoran and Zhang, Ge and Sha, Lei and Zhang, Yichang and Ren, Xuancheng and Liu, Tianyu and Chang, Baobao},
  booktitle = {International Conference on Learning Representations},
  year = {2025},
  volume = {2025},
  pages = {100540--100569},
  url = {https://proceedings.iclr.cc/paper_files/paper/2025/hash/f9e1e8b56c7e363985ebeb0e9dd1a85c-Abstract-Conference.html},
}

@inproceedings{luong2025imobench,
  title = {Towards Robust Mathematical Reasoning},
  author = {Thang Luong and Dawsen Hwang and Hoang H Nguyen and Golnaz Ghiasi and Yuri Chervonyi and Insuk Seo and Junsu Kim and Garrett Bingham and Jonathan Lee and Swaroop Mishra and Alex Zhai and Huiyi Hu and Henryk Michalewski and Jimin Kim and Jeonghyun Ahn and Junhwi Bae and Xingyou Song and Trieu Hoang Trinh and Quoc V Le and Junehyuk Jung},
  booktitle = {Proceedings of the 2025 Conference on Empirical Methods in Natural Language Processing},
  pages = {35418--35442},
  year = {2025},
  publisher = {Association for Computational Linguistics},
  doi = {10.18653/v1/2025.emnlp-main.1794},
  url = {https://aclanthology.org/2025.emnlp-main.1794/}
}

@article{gao2025closing,
  title = {Closing the Data Loop: Using {OpenDataArena} to Engineer Superior Training Datasets},
  author = {Gao, Xin and Wang, Xiaoyang and Zhu, Yun and Cai, Mengzhang and He, Conghui and Wu, Lijun},
  journal = {arXiv preprint arXiv:2601.09733},
  year = {2025},
  url = {https://arxiv.org/abs/2601.09733}
}

@article{cai2025opendataarena,
  title = {{OpenDataArena}: A Fair and Open Arena for Benchmarking Post-Training Dataset Value},
  author = {Cai, Mengzhang and Gao, Xin and Li, Yu and Lin, Honglin and Liu, Zheng and Pan, Zhuoshi and Pei, Qizhi and Shang, Xiaoran and Sun, Mengyuan and Tang, Zinan and others},
  journal = {arXiv preprint arXiv:2512.14051},
  year = {2025},
  url = {https://arxiv.org/abs/2512.14051}
}

@misc{deepseekai2026deepseekv4,
  title = {{DeepSeek-V4}: Towards Highly Efficient Million-Token Context Intelligence},
  author = {{DeepSeek-AI}},
  year = {2026},
  eprint = {2606.19348},
  archivePrefix = {arXiv},
  primaryClass = {cs.CL},
  doi = {10.48550/arXiv.2606.19348},
  url = {https://arxiv.org/abs/2606.19348}
}

@misc{yang2025qwen3,
  title = {{Qwen3} Technical Report},
  author = {An Yang and Anfeng Li and Baosong Yang and Beichen Zhang and Binyuan Hui and Bo Zheng and Bowen Yu and Chang Gao and Chengen Huang and Chenxu Lv and Chujie Zheng and Dayiheng Liu and Fan Zhou and Fei Huang and Feng Hu and Hao Ge and Haoran Wei and Huan Lin and Jialong Tang and Jian Yang and Jianhong Tu and Jianwei Zhang and Jianxin Yang and Jiaxi Yang and Jing Zhou and Jingren Zhou and Junyang Lin and Kai Dang and Keqin Bao and Kexin Yang and Le Yu and Lianghao Deng and Mei Li and Mingfeng Xue and Mingze Li and Pei Zhang and Peng Wang and Qin Zhu and Rui Men and Ruize Gao and Shixuan Liu and Shuang Luo and Tianhao Li and Tianyi Tang and Wenbiao Yin and Xingzhang Ren and Xinyu Wang and Xinyu Zhang and Xuancheng Ren and Yang Fan and Yang Su and Yichang Zhang and Yinger Zhang and Yu Wan and Yuqiong Liu and Zekun Wang and Zeyu Cui and Zhenru Zhang and Zhipeng Zhou and Zihan Qiu},
  year = {2025},
  eprint = {2505.09388},
  archivePrefix = {arXiv},
  primaryClass = {cs.CL},
  doi = {10.48550/arXiv.2505.09388},
  url = {https://arxiv.org/abs/2505.09388}
}

@misc{du2025nemotronmath,
  title = {{Nemotron-Math}: Efficient Long-Context Distillation of Mathematical Reasoning from Multi-Mode Supervision},
  author = {Wei Du and Shubham Toshniwal and Branislav Kisacanin and Sadegh Mahdavi and Ivan Moshkov and George Armstrong and Stephen Ge and Edgar Minasyan and Feng Chen and Igor Gitman},
  year = {2025},
  eprint = {2512.15489},
  archivePrefix = {arXiv},
  primaryClass = {cs.AI},
  doi = {10.48550/arXiv.2512.15489},
  url = {https://arxiv.org/abs/2512.15489}
}

@misc{maa2025aime,
  title = {2024--25 {AIME} Thresholds Are Available},
  author = {{MAA}},
  year = {2025},
  url = {https://maa.org/news/aime-thresholds-are-available/},
  note = {Updated January 6, 2025}
}

@misc{maa2026aime,
  title = {American Invitational Mathematics Examination ({AIME})},
  author = {{MAA}},
  year = {2026},
  url = {https://maa.org/maa-invitational-competitions/},
  note = {Mathematical Association of America. Official competition resource, accessed October 3, 2026}
}

@misc{hmmt2025november,
  title = {{HMMT} November 2025: Problems and Solutions},
  author = {{HMMT}},
  year = {2025},
  url = {https://www.hmmt.org/www/archive/291}
}

@misc{hmmt2026february,
  title = {{HMMT} February 2026: Problems and Solutions},
  author = {{HMMT}},
  year = {2026},
  url = {https://www.hmmt.org/www/archive/292}
}

@misc{jackson2025aaomniscience,
  title = {{AA-Omniscience}: Evaluating Cross-Domain Knowledge Reliability in Large Language Models},
  author = {Declan Jackson and William Keating and George Cameron and Micah Hill-Smith},
  year = {2025},
  eprint = {2511.13029},
  archivePrefix = {arXiv},
  url = {https://arxiv.org/abs/2511.13029}
}

@article{zhao2024swift,
  title = {{SWIFT}: A Scalable Lightweight Infrastructure for Fine-Tuning},
  author = {Yuze Zhao and Jintao Huang and Jinghan Hu and Xingjun Wang and Yunlin Mao and Daoze Zhang and Zeyinzi Jiang and Zhikai Wu and Baole Ai and Ang Wang and Wenmeng Zhou and Yingda Chen},
  journal = {Proceedings of the AAAI Conference on Artificial Intelligence},
  year = {2025},
  volume = {39},
  number = {28},
  pages = {29733--29735},
  publisher = {Association for the Advancement of Artificial Intelligence (AAAI)},
  doi = {10.1609/aaai.v39i28.35383},
  url = {https://ojs.aaai.org/index.php/AAAI/article/view/35383},
}

@misc{opencompass2023,
  title = {{OpenCompass}: A Universal Evaluation Platform for Foundation Models},
  author = {{OpenCompass Contributors}},
  year = {2023},
  url = {https://github.com/open-compass/opencompass}
}

@misc{yang2024qwen25math,
  title = {{Qwen2.5-Math} Technical Report: Toward Mathematical Expert Model via Self-Improvement},
  author = {An Yang and Beichen Zhang and Binyuan Hui and Bofei Gao and Bowen Yu and Chengpeng Li and Dayiheng Liu and Jianhong Tu and Jingren Zhou and Junyang Lin and Keming Lu and Mingfeng Xue and Runji Lin and Tianyu Liu and Xingzhang Ren and Zhenru Zhang},
  year = {2024},
  eprint = {2409.12122},
  archivePrefix = {arXiv},
  primaryClass = {cs.CL},
  doi = {10.48550/arXiv.2409.12122},
  url = {https://arxiv.org/abs/2409.12122}
}

@misc{nvidia2026nemotronmathv4,
  title = {{Nemotron-SFT-Math-v4}},
  author = {{NVIDIA}},
  year = {2026},
  url = {https://huggingface.co/datasets/nvidia/Nemotron-SFT-Math-v4},
  note = {Hugging Face dataset card}
}

@misc{nvidia2026nemotronsciencev2,
  title = {{Nemotron-SFT-Science-v2}},
  author = {{NVIDIA}},
  year = {2026},
  url = {https://huggingface.co/datasets/nvidia/Nemotron-SFT-Science-v2},
  note = {Hugging Face dataset card}
}

@misc{chen2021evaluating,
  title = {Evaluating Large Language Models Trained on Code},
  author = {Mark Chen and Jerry Tworek and Heewoo Jun and Qiming Yuan and Henrique Ponde de Oliveira Pinto and Jared Kaplan and Harri Edwards and Yuri Burda and Nicholas Joseph and Greg Brockman and Alex Ray and Raul Puri and Gretchen Krueger and Michael Petrov and Heidy Khlaaf and Girish Sastry and Pamela Mishkin and Brooke Chan and Scott Gray and Nick Ryder and Mikhail Pavlov and Alethea Power and Lukasz Kaiser and Mohammad Bavarian and Clemens Winter and Philippe Tillet and Felipe Petroski Such and Dave Cummings and Matthias Plappert and Fotios Chantzis and Elizabeth Barnes and Ariel Herbert-Voss and William Hebgen Guss and Alex Nichol and Alex Paino and Nikolas Tezak and Jie Tang and Igor Babuschkin and Suchir Balaji and Shantanu Jain and William Saunders and Christopher Hesse and Andrew N. Carr and Jan Leike and Josh Achiam and Vedant Misra and Evan Morikawa and Alec Radford and Matthew Knight and Miles Brundage and Mira Murati and Katie Mayer and Peter Welinder and Bob McGrew and Dario Amodei and Sam McCandlish and Ilya Sutskever and Wojciech Zaremba},
  year = {2021},
  eprint = {2107.03374},
  archivePrefix = {arXiv},
  primaryClass = {cs.LG},
  url = {https://arxiv.org/abs/2107.03374}
}

@misc{kimiteam2026kimik3,
  title = {{Kimi K3}: Open Frontier Intelligence},
  author = {{Kimi Team} and Tongtong Bai and Yifan Bai and Yiping Bao and M. C. and Jianfeng Cai and Xinyuan Cai and Peizhou Cao and Yuxuan Cao and Ziwei Chai and Y. Charles and H. S. Che and Guanduo Chen and Guangyu Chen and Guanzheng Chen and Huarong Chen and Jia Chen and Jianlong Chen and Jun Chen and Kexin Chen and Peng Chen and Ruijue Chen and Wentao Chen and Xin Chen and Yang Chen and Yanru Chen and Yifei Chen and Yingjiang Chen and Yuankun Chen and Yujie Chen and Yutian Chen and Zhirong Chen and Dazhi Cheng and Yean Cheng and Jialei Cui and Jingbing Cui and Anqi Dai and Jiaqi Deng and Hao Ding and Rui Ding and Shaofeng Ding and Mengfan Dong and Mengnan Dong and Yuhao Dong and Yuxin Dong and Angang Du and Chenzhuang Du and Dikang Du and Jusen Du and Yulun Du and Yu Fan and Jing Feng and Qiulin Feng and Yichen Feng and Kelin Fu and Qiang Fu and Fuxuan Gao and Hongcheng Gao and Jingyue Gao and Tong Gao and Weijia Gao and Shangyi Geng and Jie Gong and Linhu Gong and Shengao Gong and Xiaochen Gong and Qizheng Gu and Yicheng Gu and Shuhao Guan and Haiqing Guo and Shiqi Guo and Xiang Guo and Zhengyan Guo and Beixi Hao and Wenxin Hao and Xiaoru Hao and Dailan He and Haotian He and Lehan He and Qi He and Weiran He and Xinran He and Xinyi He and Yibo He and Yunjia He and Chao Hong and Tiange Hong and Hao Hu and Jiaxi Hu and Ruikun Hu and Weiming Hu and Yangyang Hu and Zhenxing Hu and Liang Hua and Jinbin Huang and Ke Huang and Ruiyuan Huang and Siying Huang and Weixiao Huang and Yan Huang and Zhengjie Huang and Zhiqi Huang and Yulong Hui and Chaobo Jia and Yutong Jiang and Zhejun Jiang and Zuoyou Jiang and Wenyi Jin and Xinyi Jin and Yu Jing and Huanjun Kong and Guokun Lai and Aidi Li and Cheng Li and Chengyuan Li and Cong Li and Fang Li and Guanyu Li and Haoyang Li and Jia Li and Junxiong Li and Lei Li and Letian Li and Lincan Li and Weihong Li and Wentao Li and Xintong Li and Yang Li and Yishen Li and Yiwei Li and Yuxiao Li and Zhaowei Li and Zhaoxi Li and Zheming Li and Zhengxiao Li and Zhiyuan Li and Jiawei Lin and Xiaohan Lin and Yibo Lin and Zichao Lin and Ziyan Lin and Bill Liu and Boxiao Liu and Chuan Liu and Liang Liu and Shaowei Liu and Shudong Liu and Shuran Liu and Tianwei Liu and Weizhou Liu and Yangyang Liu and Yanming Liu and Yibo Liu and Yipeng Liu and Zhengying Liu and Zhiheng Liu and Enzhe Lu and Haoyu Lu and Linqiang Lu and Tingzhan Lu and Zhiyuan Lu and Aotian Luo and G. Luo and Junyu Luo and Yifan Luo and B. Lyu and Wenzhou Lyu and Shaoguang Mao and Yuan Mei and Xin Men and Minqing Ni and Yixuan Niu and Siyuan Pan and Shujun Peng and Zhangyang Qi and Ruoyu Qin and ZeChao Qin and Zeyu Qin and Haiquan Qiu and Jianxin Qiu and Jiezhong Qiu and Bowen Qu and Yuhao Qu and Zeyu Shang and Youbo Shao and Han Shen and Jincheng Shi and Juanfeng Shi and Lidong Shi and Shengyuan Shi and Wingchun Siu and Pengwei Song and Xiaoxi Song and Jianlin Su and Yunfeng Su and Zhaochen Su and Lin Sui and Jingsong Sun and Junyao Sun and Shaoning Sun and Shuzhe Sun and Tongyu Sun and Yujun Sun and Yunpeng Tai and Chuning Tang and Heyi Tang and Sirui Tang and Zecheng Tang and Chaoran Tian and Rongpeng Tian and Yu Tian and Wei Tu and Chensi Wang and Chuang Wang and Chunjie Wang and Dinglu Wang and Feng Wang and Hailong Wang and Haiming Wang and Hao Wang and Hao Wang and Huaqing Wang and Hui Wang and Jiayi Wang and Jinglong Wang and Jinhong Wang and Jiuzheng Wang and Linian Wang and Shaobo Wang and Shenzhi Wang and Shuyi Wang and Si Wang and Siyuan Wang and Tianfu Wang and Wenjue Wang and Xingran Wang and Xinmei Wang and Xinyuan Wang and Xusheng Wang and Yalin Wang and Yangkun Wang and Yao Wang and Yaoyu Wang and Yejie Wang and Yiqin Wang and Yucheng Wang and Yuzhi Wang and Zhaoji Wang and Zhaowei Wang and Zhengtao Wang and Zhenhao Wang and Zhongsheng Wang and Zifan Wang and Chu Wei and Ming Wei and Shouxin Wei and Zichen Wen and Fan Wu and Haoning Wu and Rucong Wu and Wenhao Wu and Xiaoxue Wu and Yingcong Wu and Yongqi Wu and Yuxin Wu and Zijian Wu and Xinglang Xian and Chenxuan Xiang and Yuye Xiang and Bocheng Xiao and Chenjun Xiao and Xin Xiao and Jin Xie and Xiaotong Xie and Yifeng Xie and Zhe Xie and Bowei Xing and Yiming Xiong and Baosheng Xu and Boyu Xu and Jiale Xu and Jianfan Xu and Jing Xu and Jinjing Xu and L. H. Xu and Qingtao Xu and Shuyao Xu and Suting Xu and Tiantian Xu and Tianxiang Xu and Weixin Xu and Xinran Xu and Yangchuan Xu and Ye Xu and Yueni Xu and Ziyao Xu and Haonan Xue and Junjie Yan and Yaoyao Yan and Fan Yang and Guangyao Yang and Hao Yang and Junwei Yang and Ruoyu Yang and Wenjie Yang and Xiaofei Yang and Xinyu Yang and Yi Yang and Yiling Yang and Ying Yang and Yuchen Yang and Zhen Yang and Zhilin Yang and Zian Yang and Zuhao Yang and Haotian Yao and Dan Ye and Haoran Ye and Wenjie Ye and Zhanbo Ye and Bohong Yin and Haoxiang Yin and Xietong Yin and Chengzhen Yu and Haozhen Yu and Longhui Yu and Shengnan Yu and Shuying Yu and Tianxiang Yu and Enming Yuan and Mengjie Yuan and Tongtian Yue and Wei Yue and Yang Yue and Dunyuan Zha and Haobing Zhan and B. H. Zhang and Dehao Zhang and Fei Zhang and Hao Zhang and Haoyuan Zhang and Huanyu Zhang and Jiapei Zhang and Jiaxuan Zhang and Jin Zhang and Kaiyi Zhang and Miaozhen Zhang and Puqi Zhang and Qinglei Zhang and Rong Zhang and Rui Zhang and Shaoshuai Zhang and Shiyi Zhang and Xiaobin Zhang and Xiaoyun Zhang and Y. Zhang and Yangkun Zhang and Ye Zhang and Yichi Zhang and Yikun Zhang and Yizhi Zhang and Yongting Zhang and Yu Zhang and Yutao Zhang and Yutong Zhang and Zheng Zhang and Zijing Zhang and Bin Zhao and Chenguang Zhao and Feifan Zhao and Jinglun Zhao and Jinxiang Zhao and Shuai Zhao and Wenshuo Zhao and Xiangyu Zhao and Xuanle Zhao and Yikai Zhao and Zijia Zhao and Haozhi Zheng and Huabin Zheng and Ruihan Zheng and Shaojie Zheng and Tengyang Zheng and Haofeng Zhong and Lei Zhong and Longguang Zhong and M. Zhou and Qiankang Zhou and Runjie Zhou and Ruozhang Zhou and Xinyu Zhou and Yiqiao Zhou and Zaida Zhou and Jinguo Zhu and Liya Zhu and Xinhao Zhu and Yangjunfeng Zhu and Yuxuan Zhu and Zhen Zhu and Chen Zhuang and Weiyu Zhuang and Xinxing Zu},
  year = {2026},
  eprint = {2607.24653},
  archivePrefix = {arXiv},
  primaryClass = {cs.CL},
  doi = {10.48550/arXiv.2607.24653},
  url = {https://arxiv.org/abs/2607.24653}
}

@inproceedings{wang2023selfconsistency,
  title = {Self-Consistency Improves Chain of Thought Reasoning in Language Models},
  author = {Xuezhi Wang and Jason Wei and Dale Schuurmans and Quoc V Le and Ed H. Chi and Sharan Narang and Aakanksha Chowdhery and Denny Zhou},
  booktitle = {International Conference on Learning Representations},
  year = {2023},
  url = {https://openreview.net/forum?id=1PL1NIMMrw},
}

@inproceedings{zhou2023leasttomost,
  title = {Least-to-Most Prompting Enables Complex Reasoning in Large Language Models},
  author = {Denny Zhou and Nathanael Sch{\"a}rli and Le Hou and Jason Wei and Nathan Scales and Xuezhi Wang and Dale Schuurmans and Claire Cui and Olivier Bousquet and Quoc V Le and Ed H. Chi},
  booktitle = {International Conference on Learning Representations},
  year = {2023},
  url = {https://openreview.net/forum?id=WZH7099tgfM},
}

@inproceedings{ho2023reasoningteachers,
  title = {Large Language Models Are Reasoning Teachers},
  author = {Ho, Namgyu and Schmid, Laura and Yun, Se-Young},
  booktitle = {Proceedings of the 61st Annual Meeting of the Association for Computational Linguistics (Volume 1: Long Papers)},
  pages = {14852--14882},
  year = {2023},
  publisher = {Association for Computational Linguistics},
  doi = {10.18653/v1/2023.acl-long.830},
  url = {https://aclanthology.org/2023.acl-long.830/}
}

@inproceedings{hsieh2023distillingstep,
  title = {Distilling Step-by-Step! Outperforming Larger Language Models with Less Training Data and Smaller Model Sizes},
  author = {Hsieh, Cheng-Yu and Li, Chun-Liang and Yeh, Chih-kuan and Nakhost, Hootan and Fujii, Yasuhisa and Ratner, Alex and Krishna, Ranjay and Lee, Chen-Yu and Pfister, Tomas},
  booktitle = {Findings of the Association for Computational Linguistics: ACL 2023},
  pages = {8003--8017},
  year = {2023},
  publisher = {Association for Computational Linguistics},
  doi = {10.18653/v1/2023.findings-acl.507},
  url = {https://aclanthology.org/2023.findings-acl.507/}
}

@inproceedings{fu2023specializing,
  title = {Specializing Smaller Language Models towards Multi-Step Reasoning},
  author = {Fu, Yao and Peng, Hao and Ou, Litu and Sabharwal, Ashish and Khot, Tushar},
  booktitle = {Proceedings of the 40th International Conference on Machine Learning},
  year = {2023},
  volume = {202},
  pages = {10421--10430},
  publisher = {PMLR},
  url = {https://proceedings.mlr.press/v202/fu23d.html},
}

@article{guo2025deepseekr1,
  title = {{DeepSeek-R1} incentivizes reasoning in {LLMs} through reinforcement learning},
  author = {Daya Guo and Dejian Yang and Haowei Zhang and Junxiao Song and Peiyi Wang and Qihao Zhu and Runxin Xu and Ruoyu Zhang and Shirong Ma and Xiao Bi and Xiaokang Zhang and Xingkai Yu and Yu Wu and Z. F. Wu and Zhibin Gou and Zhihong Shao and Zhuoshu Li and Ziyi Gao and Aixin Liu and Bing Xue and Bingxuan Wang and Bochao Wu and Bei Feng and Chengda Lu and Chenggang Zhao and Chengqi Deng and Chong Ruan and Damai Dai and Deli Chen and Dongjie Ji and Erhang Li and Fangyun Lin and Fucong Dai and Fuli Luo and Guangbo Hao and Guanting Chen and Guowei Li and H. Zhang and Hanwei Xu and Honghui Ding and Huazuo Gao and Hui Qu and Hui Li and Jianzhong Guo and Jiashi Li and Jingchang Chen and Jingyang Yuan and Jinhao Tu and Junjie Qiu and Junlong Li and J. L. Cai and Jiaqi Ni and Jian Liang and Jin Chen and Kai Dong and Kai Hu and Kaichao You and Kaige Gao and Kang Guan and Kexin Huang and Kuai Yu and Lean Wang and Lecong Zhang and Liang Zhao and Litong Wang and Liyue Zhang and Lei Xu and Leyi Xia and Mingchuan Zhang and Minghua Zhang and Minghui Tang and Mingxu Zhou and Meng Li and Miaojun Wang and Mingming Li and Ning Tian and Panpan Huang and Peng Zhang and Qiancheng Wang and Qinyu Chen and Qiushi Du and Ruiqi Ge and Ruisong Zhang and Ruizhe Pan and Runji Wang and R. J. Chen and R. L. Jin and Ruyi Chen and Shanghao Lu and Shangyan Zhou and Shanhuang Chen and Shengfeng Ye and Shiyu Wang and Shuiping Yu and Shunfeng Zhou and Shuting Pan and S. S. Li and Shuang Zhou and Shaoqing Wu and Tao Yun and Tian Pei and Tianyu Sun and T. Wang and Wangding Zeng and Wen Liu and Wenfeng Liang and Wenjun Gao and Wenqin Yu and Wentao Zhang and W. L. Xiao and Wei An and Xiaodong Liu and Xiaohan Wang and Xiaokang Chen and Xiaotao Nie and Xin Cheng and Xin Liu and Xin Xie and Xingchao Liu and Xinyu Yang and Xinyuan Li and Xuecheng Su and Xuheng Lin and X. Q. Li and Xiangyue Jin and Xiaojin Shen and Xiaosha Chen and Xiaowen Sun and Xiaoxiang Wang and Xinnan Song and Xinyi Zhou and Xianzu Wang and Xinxia Shan and Y. K. Li and Y. Q. Wang and Y. X. Wei and Yang Zhang and Yanhong Xu and Yao Li and Yao Zhao and Yaofeng Sun and Yaohui Wang and Yi Yu and Yichao Zhang and Yifan Shi and Yiliang Xiong and Ying He and Yishi Piao and Yisong Wang and Yixuan Tan and Yiyang Ma and Yiyuan Liu and Yongqiang Guo and Yuan Ou and Yuduan Wang and Yue Gong and Yuheng Zou and Yujia He and Yunfan Xiong and Yuxiang Luo and Yuxiang You and Yuxuan Liu and Yuyang Zhou and Y. X. Zhu and Yanping Huang and Yaohui Li and Yi Zheng and Yuchen Zhu and Yunxian Ma and Ying Tang and Yukun Zha and Yuting Yan and Z. Z. Ren and Zehui Ren and Zhangli Sha and Zhe Fu and Zhean Xu and Zhenda Xie and Zhengyan Zhang and Zhewen Hao and Zhicheng Ma and Zhigang Yan and Zhiyu Wu and Zihui Gu and Zijia Zhu and Zijun Liu and Zilin Li and Ziwei Xie and Ziyang Song and Zizheng Pan and Zhen Huang and Zhipeng Xu and Zhongyu Zhang and Zhen Zhang},
  journal = {Nature},
  volume = {645},
  number = {8081},
  pages = {633--638},
  year = {2025},
  doi = {10.1038/s41586-025-09422-z},
  url = {https://www.nature.com/articles/s41586-025-09422-z}
}

@inproceedings{ludziejewski2025jointmoe,
  title = {Joint {MoE} Scaling Laws: Mixture of Experts Can Be Memory Efficient},
  author = {Ludziejewski, Jan and Pi\'{o}ro, Maciej and Krajewski, Jakub and Stefaniak, Maciej and Krutul, Micha{\l} and Ma{\l}a\'{s}nicki, Jan and Cygan, Marek and Sankowski, Piotr and Adamczewski, Kamil and Mi{\l}o\'{s}, Piotr and Jaszczur, Sebastian},
  booktitle = {Proceedings of the 42nd International Conference on Machine Learning},
  volume = {267},
  series = {Proceedings of Machine Learning Research},
  pages = {41056--41073},
  year = {2025},
  publisher = {PMLR},
  url = {https://proceedings.mlr.press/v267/ludziejewski25a.html}
}

@inproceedings{pillutla2021mauve,
  title = {{MAUVE}: Measuring the Gap Between Neural Text and Human Text using Divergence Frontiers},
  author = {Krishna Pillutla and Swabha Swayamdipta and Rowan Zellers and John Thickstun and Sean Welleck and Yejin Choi and Zaid Harchaoui},
  booktitle = {Advances in Neural Information Processing Systems},
  volume = {34},
  pages = {4816--4828},
  year = {2021},
  url = {https://proceedings.neurips.cc/paper/2021/hash/260c2432a0eecc28ce03c10dadc078a4-Abstract.html}
}

@article{gretton2012kernel,
  title = {A Kernel Two-Sample Test},
  author = {Arthur Gretton and Karsten M. Borgwardt and Malte J. Rasch and Bernhard Sch{\"o}lkopf and Alexander Smola},
  journal = {Journal of Machine Learning Research},
  volume = {13},
  number = {25},
  pages = {723--773},
  year = {2012},
  url = {https://jmlr.org/papers/v13/gretton12a.html}
}

@article{zhang2025qwen3embedding,
  title = {{Qwen3 Embedding}: Advancing Text Embedding and Reranking Through Foundation Models},
  author = {Yanzhao Zhang and Mingxin Li and Dingkun Long and Xin Zhang and Huan Lin and Baosong Yang and Pengjun Xie and An Yang and Dayiheng Liu and Junyang Lin and Fei Huang and Jingren Zhou},
  journal = {arXiv preprint arXiv:2506.05176},
  year = {2025},
  url = {https://arxiv.org/abs/2506.05176}
}

@inproceedings{li2023best,
  title = {The Best of Both Worlds: Combining Human and Machine Translations for Multilingual Semantic Parsing with Active Learning},
  author = {Li, Zhuang and Qu, Lizhen and Cohen, Philip and Tumuluri, Raj and Haffari, Gholamreza},
  booktitle = {Proceedings of the 61st Annual Meeting of the Association for Computational Linguistics (Volume 1: Long Papers)},
  pages = {9511--9528},
  year = {2023},
  publisher = {Association for Computational Linguistics},
  doi = {10.18653/v1/2023.acl-long.529},
  url = {https://aclanthology.org/2023.acl-long.529/}
}

@misc{mitsuzawa2025mmdflagger,
  title = {{MMD-Flagger}: Leveraging Maximum Mean Discrepancy to Detect Hallucinations},
  author = {Kensuke Mitsuzawa and Damien Garreau},
  year = {2025},
  eprint = {2506.01367v1},
  archivePrefix = {arXiv},
  primaryClass = {cs.CL},
  url = {https://arxiv.org/abs/2506.01367v1}
}

@techreport{openai2026gpt6systemcard,
  title = {{GPT-6 Astra System Card}},
  author = {{OpenAI}},
  institution = {OpenAI},
  year = {2026},
  url = {https://deploymentsafety.openai.com/gpt-6-astra/sec:appendix-sol-luna},
  note = {Appendix: {GPT-6 Sol}, {GPT-6 Luna}}
}

\appendix
\section{Experimental details}
\label{sec:experimental_details}

\subsection{Datasets}

The following datasets are used for translator training, student distillation, and transfer experiments.

\textbf{\texttt{nvidia/\allowbreak{}Nemotron-SFT-Math-v4}} contains mathematical problems from \texttt{nvidia/\allowbreak{}Nemotron-Math-v2}, sourced from Art of Problem Solving and Math StackExchange/MathOverflow.
Solutions are generated by \texttt{DeepSeek-V4-Pro} in High inference mode and kept only when their final answers match the references.
The full release contains $545,431$ records, including $285,516$ in the \textbf{\texttt{cot}} subset~\citep{nvidia2026nemotronmathv4}.
We use all \texttt{cot} problems to generate student training data and a $1/8$ subset (approximately \texttt{35.7k} paired examples) to train the main translator.

\textbf{\texttt{OpenDataArena/\allowbreak{}ODA-Math-460k}} combines open mathematics corpora curated through deduplication, benchmark decontamination, LLM-based filtering, and verifier-backed response distillation~\citep{gao2025closing,cai2025opendataarena}.
The released dataset contains $459,646$ records, all of which we use to test transfer to a new math dataset.

\textbf{\texttt{nvidia/\allowbreak{}Nemotron-SFT-Science-v2}} contains scientific reasoning problems and model-generated solutions~\citep{nvidia2026nemotronsciencev2}.
Its \textbf{\texttt{vendor}} subset contains $221,598$ non-synthetic, graduate and research-level open-ended problems in physics, chemistry, and biology.
We use this subset to generate teacher responses for the STEM transfer experiment and for its compute budget.

\subsection{Models}

Teacher responses are generated with the following models.

\textbf{\texttt{deepseek-ai/\allowbreak{}DeepSeek-V4-Pro}} is a mixture-of-experts (MoE) language model with 1.6T parameters (49B activated).
It is our main teacher for RCG generation, and its verified FCG responses provide the translator training targets.

\textbf{\texttt{moonshotai/\allowbreak{}Kimi-K3}} is an open-weight multimodal MoE model with 2.8T parameters (104B activated).
We use it as an alternative teacher on \texttt{Nemotron-SFT-Math-v4} to test transfer across teacher models.

Our student models come from the \texttt{Qwen3} and \texttt{Qwen2.5-Math} families.

\textbf{\texttt{Qwen3}} is a family of dense and MoE language models.
We use \texttt{Qwen3-4B-Thinking-2507} as the main translator, with \texttt{Qwen3-\{1.7B, 8B\}} as alternatives for the size analysis.
We use \texttt{Qwen3-\{4B-Base, 4B-Thinking-2507, 8B-Base, 8B\}} as students.

\textbf{\texttt{Qwen2.5-Math}} is a family of language models specialized in mathematics.
We use \texttt{Qwen2.5-Math-\{7B, 7B-Instruct\}} as students to evaluate distillation on a different model family.

\subsection{Benchmarks}

Competition benchmarks evaluate mathematical reasoning, and STEM benchmarks assess domain transfer.

\textbf{\texttt{AIME 2025}} and \textbf{\texttt{AIME 2026}} are problem sets from the American Invitational Mathematics Examination.
They assess mathematical problem solving with integer answers between $0$ and $999$.

\textbf{\texttt{HMMT November 2025}} and \textbf{\texttt{HMMT February 2026}} contain problems from the Harvard--MIT Mathematics Tournament.
They assess mathematical problem solving with short answers across topics such as algebra, geometry, and combinatorics.

\textbf{\texttt{Omni-MATH}} contains $4,428$ olympiad-level mathematical problems spanning diverse topics and difficulty levels.

\textbf{\texttt{IMO-AnswerBench}} comprises $400$ olympiad-level mathematics problems with verifiable short answers.

For STEM transfer, we use the STEM subset of \texttt{OlympiadBench}, along with \texttt{SciBench} and \texttt{AA-Omniscience}.

\textbf{\texttt{OlympiadBench}} contains mathematics and physics problems drawn from olympiads and entrance examinations.

\textbf{\texttt{SciBench}} contains college-level, open-ended problems in mathematics, physics, and chemistry, requiring scientific knowledge, multistep reasoning, and numerical computation.

\textbf{\texttt{AA-Omniscience}} evaluates factual knowledge and calibration using $6,000$ questions across 42 topics in six domains.
We evaluate only on the 600 publicly available questions.

\subsection{Hyperparameters}

The random seed for model training is set to 42, and all experiments are conducted on NVIDIA H800 GPUs.

\begin{table*}[!htb]
\centering
\begin{minipage}[t]{0.32\textwidth}
\centering
\resizebox{\linewidth}{!}{%
\begin{tabular}{@{}l>{\centering\arraybackslash}p{7em}@{}}
\toprule
\textbf{Hyperparameter} & \textbf{Value} \\
\midrule
\rowcolor{antblue!15}\multicolumn{2}{c}{\texttt{Qwen3-1.7B}} \\
\texttt{lr} & \texttt{5e-5} \\
\texttt{num\_train\_epochs} & \texttt{3} \\
\texttt{lr\_warmup\_fraction} & \texttt{0.1} \\
\texttt{weight\_decay} & \texttt{0.1} \\
\texttt{clip\_grad} & \texttt{1.0} \\
\texttt{micro\_batch\_size} & \texttt{2} \\
\texttt{global\_batch\_size} & \texttt{32} \\
\texttt{max\_length} & \texttt{65536} \\
\texttt{template} & \texttt{qwen3} \\
\texttt{packing} & \texttt{true} \\
\texttt{padding\_free} & \texttt{true} \\
\texttt{bf16} & \texttt{true} \\
\texttt{attention\_backend} & \texttt{flash\_3} \\
\texttt{context\_parallel\_size} & \texttt{2} \\
\texttt{tensor\_model\_parallel\_size} & \texttt{1} \\
\bottomrule
\end{tabular}%
}
\end{minipage}\hfill
\begin{minipage}[t]{0.32\textwidth}
\centering
\resizebox{\linewidth}{!}{%
\begin{tabular}{@{}l>{\centering\arraybackslash}p{7em}@{}}
\toprule
\textbf{Hyperparameter} & \textbf{Value} \\
\midrule
\rowcolor{antblue!15}\multicolumn{2}{c}{\texttt{Qwen3-4B-Thinking-2507}} \\
\texttt{lr} & \texttt{5e-5} \\
\texttt{num\_train\_epochs} & \texttt{3} \\
\texttt{lr\_warmup\_fraction} & \texttt{0.1} \\
\texttt{weight\_decay} & \texttt{0.1} \\
\texttt{clip\_grad} & \texttt{1.0} \\
\texttt{micro\_batch\_size} & \texttt{2} \\
\texttt{global\_batch\_size} & \texttt{32} \\
\texttt{max\_length} & \texttt{65536} \\
\texttt{template} & \texttt{qwen3\_thinking} \\
\texttt{packing} & \texttt{true} \\
\texttt{padding\_free} & \texttt{true} \\
\texttt{bf16} & \texttt{true} \\
\texttt{attention\_backend} & \texttt{flash\_3} \\
\texttt{context\_parallel\_size} & \texttt{4} \\
\texttt{tensor\_model\_parallel\_size} & \texttt{1} \\
\bottomrule
\end{tabular}%
}
\end{minipage}\hfill
\begin{minipage}[t]{0.32\textwidth}
\centering
\resizebox{\linewidth}{!}{%
\begin{tabular}{@{}l>{\centering\arraybackslash}p{7em}@{}}
\toprule
\textbf{Hyperparameter} & \textbf{Value} \\
\midrule
\rowcolor{antblue!15}\multicolumn{2}{c}{\texttt{Qwen3-8B}} \\
\texttt{lr} & \texttt{5e-5} \\
\texttt{num\_train\_epochs} & \texttt{3} \\
\texttt{lr\_warmup\_fraction} & \texttt{0.1} \\
\texttt{weight\_decay} & \texttt{0.1} \\
\texttt{clip\_grad} & \texttt{1.0} \\
\texttt{micro\_batch\_size} & \texttt{1} \\
\texttt{global\_batch\_size} & \texttt{32} \\
\texttt{max\_length} & \texttt{65536} \\
\texttt{template} & \texttt{qwen3} \\
\texttt{packing} & \texttt{true} \\
\texttt{padding\_free} & \texttt{true} \\
\texttt{bf16} & \texttt{true} \\
\texttt{attention\_backend} & \texttt{flash\_3} \\
\texttt{context\_parallel\_size} & \texttt{2} \\
\texttt{tensor\_model\_parallel\_size} & \texttt{2} \\
\bottomrule
\end{tabular}%
}
\end{minipage}
\caption{Training hyperparameters for the translator.}
\label{tab:hyper_translator}
\end{table*}

\begin{table*}[!htb]
\centering
\begin{minipage}[t]{0.32\textwidth}
\centering
\resizebox{\linewidth}{!}{%
\begin{tabular}{@{}l>{\centering\arraybackslash}p{8em}@{}}
\toprule
\textbf{Hyperparameter} & \textbf{Value} \\
\midrule
\rowcolor{antblue!15}\multicolumn{2}{c}{\texttt{Qwen2.5-Math-\{7B, 7B-Instruct\}}} \\
\texttt{lr} & \texttt{5e-5} \\
\texttt{num\_train\_epochs} & \texttt{3} \\
\texttt{lr\_warmup\_fraction} & \texttt{0.1} \\
\texttt{weight\_decay} & \texttt{0.1} \\
\texttt{clip\_grad} & \texttt{1.0} \\
\texttt{micro\_batch\_size} & \texttt{1} \\
\texttt{global\_batch\_size} & \texttt{32} \\
\texttt{max\_length} & \texttt{32768} \\
\texttt{packing} & \texttt{true} \\
\texttt{padding\_free} & \texttt{true} \\
\texttt{bf16} & \texttt{true} \\
\texttt{attention\_backend} & \texttt{flash\_3} \\
\texttt{context\_parallel\_size} & \texttt{2} \\
\texttt{tensor\_model\_parallel\_size} & \texttt{1} \\
\midrule
\rowcolor{antblue!7}\multicolumn{2}{c}{\texttt{template}} \\
\texttt{7B} & \texttt{default} \\
\texttt{7B-Instruct} & \texttt{qwen2\_5\_math} \\
\bottomrule
\end{tabular}%
}
\end{minipage}\hfill
\begin{minipage}[t]{0.32\textwidth}
\centering
\resizebox{\linewidth}{!}{%
\begin{tabular}{@{}l>{\centering\arraybackslash}p{8em}@{}}
\toprule
\textbf{Hyperparameter} & \textbf{Value} \\
\midrule
\rowcolor{antblue!15}\multicolumn{2}{c}{\texttt{Qwen3-\{4B-Base, 4B-Thinking-2507\}}} \\
\texttt{lr} & \texttt{5e-5} \\
\texttt{num\_train\_epochs} & \texttt{3} \\
\texttt{lr\_warmup\_fraction} & \texttt{0.1} \\
\texttt{weight\_decay} & \texttt{0.1} \\
\texttt{clip\_grad} & \texttt{1.0} \\
\texttt{micro\_batch\_size} & \texttt{2} \\
\texttt{global\_batch\_size} & \texttt{32} \\
\texttt{max\_length} & \texttt{32768} \\
\texttt{packing} & \texttt{true} \\
\texttt{padding\_free} & \texttt{true} \\
\texttt{bf16} & \texttt{true} \\
\texttt{attention\_backend} & \texttt{flash\_3} \\
\texttt{context\_parallel\_size} & \texttt{1} \\
\texttt{tensor\_model\_parallel\_size} & \texttt{1} \\
\midrule
\rowcolor{antblue!7}\multicolumn{2}{c}{\texttt{template}} \\
\texttt{4B-Base} & \texttt{default} \\
\texttt{4B-Thinking-2507} & \texttt{qwen3\_thinking} \\
\bottomrule
\end{tabular}%
}
\end{minipage}\hfill
\begin{minipage}[t]{0.32\textwidth}
\centering
\resizebox{\linewidth}{!}{%
\begin{tabular}{@{}l>{\centering\arraybackslash}p{8em}@{}}
\toprule
\textbf{Hyperparameter} & \textbf{Value} \\
\midrule
\rowcolor{antblue!15}\multicolumn{2}{c}{\texttt{Qwen3-\{8B-Base, 8B\}}} \\
\texttt{lr} & \texttt{5e-5} \\
\texttt{num\_train\_epochs} & \texttt{3} \\
\texttt{lr\_warmup\_fraction} & \texttt{0.1} \\
\texttt{weight\_decay} & \texttt{0.1} \\
\texttt{clip\_grad} & \texttt{1.0} \\
\texttt{micro\_batch\_size} & \texttt{1} \\
\texttt{global\_batch\_size} & \texttt{32} \\
\texttt{max\_length} & \texttt{32768} \\
\texttt{packing} & \texttt{true} \\
\texttt{padding\_free} & \texttt{true} \\
\texttt{bf16} & \texttt{true} \\
\texttt{attention\_backend} & \texttt{flash\_3} \\
\texttt{context\_parallel\_size} & \texttt{2} \\
\texttt{tensor\_model\_parallel\_size} & \texttt{1} \\
\midrule
\rowcolor{antblue!7}\multicolumn{2}{c}{\texttt{template}} \\
\texttt{8B-Base} & \texttt{default} \\
\texttt{8B} & \texttt{qwen3} \\
\bottomrule
\end{tabular}%
}
\end{minipage}
\caption{Training hyperparameters for the student models.}
\label{tab:hyper_student}
\end{table*}

\begin{table*}[!htb]
\centering
\begin{minipage}[t]{0.49\textwidth}
\centering
\begin{tabular}{@{}p{12.5em}>{\centering\arraybackslash}p{7em}@{}}
\toprule
\textbf{Hyperparameter} & \textbf{Value} \\
\midrule
\rowcolor{antblue!15}\multicolumn{2}{c}{\texttt{teacher model}} \\
\texttt{temperature} & \texttt{1.0} \\
\texttt{reasoning\_effort} & \texttt{high} \\
\texttt{max\_out\_token} & \texttt{32768} \\
\rowcolor{antblue!7}\multicolumn{2}{c}{\texttt{top\_p}} \\
\texttt{DeepSeek-V4-Pro} & \texttt{1.0} \\
\texttt{Kimi-K3} & \texttt{0.95} \\
\bottomrule
\end{tabular}
\end{minipage}\hfill
\begin{minipage}[t]{0.49\textwidth}
\centering
\begin{tabular}{@{}p{12.5em}>{\centering\arraybackslash}p{7em}@{}}
\toprule
\textbf{Hyperparameter} & \textbf{Value} \\
\midrule
\rowcolor{antblue!15}\multicolumn{2}{c}{\texttt{translator}} \\
\texttt{server} & \texttt{sglang} \\
\texttt{temperature} & \texttt{0.6} \\
\texttt{top\_p} & \texttt{0.95} \\
\texttt{max\_out\_token} & \texttt{32768} \\
\texttt{top\_k} & \texttt{20} \\
\texttt{min\_p} & \texttt{0} \\
\bottomrule
\end{tabular}
\end{minipage}
\caption{Inference hyperparameters for teacher models and translators.}
\label{tab:hyper_infer}
\end{table*}

\paragraph{Training.}
\autoref{tab:hyper_translator} and~\autoref{tab:hyper_student} list translator and student training hyperparameters.
For models with a shorter native context window, we enable \texttt{rope\_scaling} in \texttt{config.json} to support the required training length: \texttt{32k} tokens for students and \texttt{64k} tokens for translators.

\paragraph{Inference.}
For teacher generation and translator inference, we use the officially recommended or default sampling settings listed in~\autoref{tab:hyper_infer}.

\section{Substantive AI assistance information}
\label{sec:ai_assistance}

We used AI tools to assist with manuscript writing, implementation and debugging of some code, and literature searches.
AI tools were not involved in developing the research ideas, methodology, experimental design, or scientific conclusions, which were developed independently by the authors.

\section{Mathematical analysis}
\label{sec:mathematical_analysis}

\method fits a distribution over complete forward responses, including termination, from paired reasoning responses, following the probabilistic sequence modeling used in machine translation~\citep{sutskever2014sequence,kim2016sequence}.
Write $c=(q,a)$, $R_c=P_{\mathrm{RCG}}(\cdot\mid q,a)$, and $T_c=P_{\mathrm{RS}}(\cdot\mid q,a)$.
Uppercase letters denote the corresponding random variables, with $C=(Q,A)$, and $\widetilde P_{\phi,c}=\widetilde P_\phi(\cdot\mid q,a)$.
For the population analysis, we assume independent candidate sampling from a question-answer distribution $\mu(c)$, independent source and target generation given $c$, independent forward attempts, and an exact answer check.
The teacher acceptance probability is positive on the candidate support; logarithms are natural, and population losses and entropies are finite.

\subsection{Pairs preserve the target law}
\label{sec:target_law_analysis}

For an attempt limit $K$ and $p=Z_{\mathcal T}(q,a)$, the probability of acceptance within $K$ attempts is $C_K(c)=1-(1-p)^K$.
With $\tau$ the first accepted attempt, summing its probabilities over these attempts gives
$\Pr(Z_\tau=z\mid\tau\le K,c)=T_c(z)$.
Thus the attempt limit changes which questions are kept, while preserving their conditional target law.
The law of accepted contexts and the population pair distribution are
\begin{align}
  \nu(c)&=\frac{\mu(c)C_K(c)}{\mathbb E_\mu C_K(C)},
  \label{eq:retained_context}\\
  \pi(c,x,z)&=\nu(c)R_c(x)T_c(z).
  \label{eq:paired_law}
\end{align}
Under $\pi$, $(C,X,Z)$ are the context, source, and accepted target; $\pi_X$ denotes the source marginal.
This construction supplies samples from verified question-only generation, even though the source is answer-conditioned.

\subsection{Likelihood enables recovery}
\label{sec:paired_identification}
\label{sec:recovery_accuracy}

Under autoregressive factorization, teacher forcing in~\autoref{eq:loss} minimizes sequence cross-entropy~\citep{kim2016sequence}.
Since the logarithmic score is strictly proper, its optimum over unrestricted conditional distributions is $\mathcal G^*(z\mid x)=\pi(z\mid x)$~\citep{gneiting2007proper}.
The population counterpart of~\autoref{eq:loss} is $\mathcal L(\phi)=-\mathbb E_\pi\log\mathcal G_\phi(Z\mid X)$; define
\begin{align}
  \epsilon_\phi&=\mathcal L(\phi)-H_\pi(Z\mid X),
  \label{eq:excess_loss}\\
  \delta_{\mathrm{src}}&=I_\pi(Z;C\mid X).
  \label{eq:source_information}
\end{align}
Here $H_\pi$ and $I_\pi$ denote conditional entropy and conditional mutual information.
The first term is the average conditional KL fitting error; the second measures target information missing from the source.
For the induced model law $\widetilde P_{\phi,c}$ in~\autoref{eq:marginal}, the log-sum inequality bounds marginal KL by joint KL~\citep{cover2006elements}.
Using $\pi(z\mid c,x)=T_c(z)$ and inserting $\pi(Z\mid X)$ gives
\begin{align}
  &\mathbb E_\nu D_{\mathrm{KL}}(T_C\Vert\widetilde P_{\phi,C})\nonumber\\
  &\quad\le\mathbb E_\pi\log\frac{T_C(Z)}{\mathcal G_\phi(Z\mid X)}\nonumber\\
  &\quad=I_\pi(Z;C\mid X)+\epsilon_\phi
   =\delta_{\mathrm{src}}+\epsilon_\phi.
  \label{eq:recovery_bound}
\end{align}

If $x$ contains enough task information to determine $T_c$, then $\delta_{\mathrm{src}}=0$.
Together with exact population fitting, this yields exact recovery for $\nu$-almost every context.
Source and target responses need not align step by step: the source contains the relevant task information, and paired likelihood learns its forward response law.
With imperfect fitting, let $\eta_\phi=\delta_{\mathrm{src}}+\epsilon_\phi$.
Pinsker's and Jensen's inequalities give
\begin{equation}
  \mathbb E_\nu\operatorname{TV}(T_C,\widetilde P_{\phi,C})
  \le\sqrt{\eta_\phi/2},
  \label{eq:recovery_tv}
\end{equation}
where $\operatorname{TV}(P,Q)=\tfrac12\sum_z|P(z)-Q(z)|$.
Thus probabilities of properties of responses differ by at most this amount on average.
Here $\epsilon_\phi$ is a population error; small training loss alone does not establish recovery on unseen pairs.

\subsection{Recovery controls student risk}
\label{sec:filter_distillation_analysis}

The inference law $\widehat{\mathcal G}_\phi$ in~\autoref{tab:hyper_infer} includes temperature adjustment and truncation.
Its discrepancy from the likelihood model is
\begin{equation}
  d_{\mathrm{dec}}=\mathbb E_{\pi_X}\operatorname{TV}
  (\mathcal G_\phi(\cdot\mid X),\widehat{\mathcal G}_\phi(\cdot\mid X)).
  \label{eq:decoder_error}
\end{equation}
For $\widehat P_{\phi,c}=\sum_xR_c(x)\widehat{\mathcal G}_\phi(\cdot\mid x)$, convexity of TV adds at most $d_{\mathrm{dec}}$ to~\autoref{eq:recovery_tv}.
Let $\widehat r_{\phi,c}=\widehat P_{\phi,c}[F(Z)=a]>0$ and $\widehat{\bar P}_{\phi,c}$ be its answer-filtered law.
Filtering preserves the TV bound: $T_c$ has only correct responses, and normalization increases the translated probability of each such response, increasing its overlap with $T_c$.

Filtering also changes the context marginal to
\begin{equation}
  \rho_\phi(c)=\frac{\mu(c)\widehat r_{\phi,c}}
    {\mathbb E_\mu\widehat r_{\phi,C}}.
  \label{eq:student_context_law}
\end{equation}
For a bounded student loss $0\le\ell_\theta(q,z)\le B$, such as clipped sequence NLL for $\mathcal S_\theta$, let $\mathcal R_{\mathrm{RS}}(\theta)$ and $\mathcal R_\phi(\theta)$ denote its expectations under $\nu T$ and $\rho_\phi\widehat{\bar P}_\phi$, respectively.
The triangle inequality gives
\begin{align}
  &|\mathcal R_{\mathrm{RS}}(\theta)-\mathcal R_\phi(\theta)|\nonumber\\
  &\quad\le B\bigl(\sqrt{\eta_\phi/2}+d_{\mathrm{dec}}
       +\operatorname{TV}(\rho_\phi,\nu)\bigr).
  \label{eq:distillation_bound}
\end{align}
The terms account for response recovery, decoding, and question coverage.
The raw student NLL in~\autoref{eq:student} is unbounded and is not covered by this inequality.
A population minimizer of $\mathcal R_\phi$ has excess target risk within the same student class at most twice this bound.
The guarantee concerns verified FCG responses; final-answer correctness does not certify causal faithfulness~\citep{lanham2023faithfulness}.
Transfer to other teachers or domains is evaluated empirically in~\autoref{sec:generalization}.

\section{Analysis details}
\label{sec:analysis_details}

\subsection{Translator and training data size}
\label{sec:translator_scale_details}

Detailed benchmark scores for the analysis in~\autoref{sec:translator_scale} are provided in~\autoref{tab:translator_scale_details}.
All settings use a \texttt{Qwen3-4B-Base} student trained on responses for the same held-out half of \texttt{nvidia/\allowbreak{}Nemotron-SFT-Math-v4}.
SUP uses untranslated responses, while the other settings use responses translated by the indicated model.

\begin{table*}[!htb]
\centering
\begin{tabular*}{\linewidth}{@{\extracolsep{\fill}}lcccccc>{\columncolor{antblue!7}[\tabcolsep][0pt]}c@{}}
\shline
\multirow{2}{*}{\textbf{Setting}} & \textbf{\texttt{AIME25}} & \textbf{\texttt{AIME26}} & \textbf{\texttt{HMMT25}} & \textbf{\texttt{HMMT26}} & \textbf{\texttt{Omni}} & \textbf{\texttt{IMO}} & \\
 & \textcolor{gray}{\texttt{pass@64\_1}} & \textcolor{gray}{\texttt{pass@64\_1}} & \textcolor{gray}{\texttt{pass@64\_1}} & \textcolor{gray}{\texttt{pass@64\_1}} & \textcolor{gray}{\texttt{pass@1\_1}} & \textcolor{gray}{\texttt{pass@8\_1}} & \multirow{-2}{*}{\textit{\textbf{Avg.}}} \\
\shline
SUP & 21.88 & 21.15 & 10.57 & 13.07 & 28.88 & 10.25 & 17.63 \\
\shline

\rowcolor{antblue!15}\multicolumn{8}{c}{\texttt{Qwen3-1.7B}} \\
\shline
$1/8$ & 10.63 & 9.53 & 3.91 & 5.02 & 23.62 & 4.91 & 9.60 \\
$1/4$ & 12.81 & 7.92 & 4.32 & 6.58 & 24.73 & 5.19 & 10.26 \\
$1/2$ & 13.39 & 11.15 & 5.78 & 8.48 & 28.00 & 6.19 & 12.17 \\
\shline

\rowcolor{antblue!15}\multicolumn{8}{c}{\texttt{Qwen3-4B-Thinking-2507}} \\
\shline
$0$ & 6.30 & 4.01 & 3.70 & 3.36 & 21.14 & 5.81 & 7.39 \\
$1/8$ & 24.64 & 27.97 & 18.02 & 18.61 & 38.50 & 12.88 & 23.44 \\
$1/4$ & 24.90 & 26.56 & 16.41 & 17.23 & 37.20 & 13.69 & 22.67 \\
$1/2$ & 24.58 & 28.28 & 17.14 & 17.99 & 37.22 & 12.22 & 22.91 \\
\shline

\rowcolor{antblue!15}\multicolumn{8}{c}{\texttt{Qwen3-8B}} \\
\shline
$1/8$ & 24.11 & 26.46 & 14.11 & 15.67 & 34.35 & 11.16 & 20.98 \\
$1/4$ & 22.92 & 24.27 & 13.80 & 15.86 & 35.28 & 11.84 & 20.66 \\
$1/2$ & 24.64 & 25.68 & 13.13 & 15.67 & 36.36 & 12.66 & 21.36 \\
\shline
\end{tabular*}
\caption{Detailed math results for the translator size and training-data analysis.
The fractions denote paired training subsets of the \texttt{cot} data in \texttt{nvidia/\allowbreak{}Nemotron-SFT-Math-v4}: $1/8$, $1/4$, and $1/2$ correspond to \texttt{35.7k}, \texttt{71.4k}, and \texttt{142.8k} examples, respectively.
The $0$ setting uses no paired training examples.}
\label{tab:translator_scale_details}
\end{table*}

\subsection{Similarity to RS responses}
\label{sec:teacher_resemblance_details}

We compare SSR, SUP, SUP-AUG, and~\method against the original verified FCG responses in \texttt{Nemotron-SFT-Math-v4}, denoted RS.
Difficulty is measured by the separate 16-attempt rejection-sampling run in~\autoref{sec:teacher_resemblance}; $>16$ denotes no success within that run.
For each difficulty group, let $t_i$ and $r_i$ be the $n$ matched RS and method responses.
We compute the four metrics separately with each feature type.

\paragraph{Features.}
Define $\operatorname{norm}(v)=v/\|v\|_2$.
For a full response $r$, the semantic and lexical features are
\[
\begin{aligned}
  f_{\mathrm{sem}}(r)&=\operatorname{norm}\bigl(E(r)\bigr),\\
  f_{\mathrm{lex}}(r)&=\operatorname{norm}\bigl(S_{128}\,\operatorname{TFIDF}(r)\bigr).
\end{aligned}
\]
Here $E$ is \texttt{Qwen3-Embedding-4B}; TF-IDF uses word, number, and math-symbol unigrams and bigrams; and $S_{128}$ is a 128-dimensional SVD projection.
The TF-IDF vocabulary and SVD projection are fitted once over all compared responses.
Below, $x_i=f(t_i)$ and $y_i=f(r_i)$.

\paragraph{MAUVE.}
PCA keeps 90\% of feature variance, and 20-cluster k-means gives histograms $p$ and $q$ for the RS and method responses.
With $m_\lambda=\lambda p+(1-\lambda)q$, the divergence curve and its area are
\[
\begin{aligned}
  a_\lambda&=e^{-5D_{\mathrm{KL}}(q\|m_\lambda)},\\
  b_\lambda&=e^{-5D_{\mathrm{KL}}(p\|m_\lambda)},\\
  \operatorname{MAUVE}&=\tfrac12\bigl[\operatorname{AUC}(a,b)+\operatorname{AUC}(b,a)\bigr].
\end{aligned}
\]
We use 25 mixture weights, include the endpoints $(0,1)$ and $(1,0)$, and average five clustering seeds.

\paragraph{MMD.}
We use the biased RBF estimate
\[
\begin{aligned}
  A_{ij}&=\kappa(x_i,x_j)+\kappa(y_i,y_j)-2\kappa(x_i,y_j),\\
  \operatorname{MMD}&=\sqrt{\frac{1}{n^2}\sum_{i,j=1}^{n}A_{ij}},
\end{aligned}
\]
where $\kappa(u,v)=\exp(-\|u-v\|_2^2/(2\sigma^2))$.
For each feature type, $\sigma^2$ is the median squared distance in a pooled sample and is fixed across methods and groups.

\paragraph{Cosine similarity.}
Since both feature types are unit normalized, the mean paired cosine similarity is
\[
  \operatorname{Cosine}=\frac{1}{n}\sum_{i=1}^{n}x_i^\top y_i.
\]

\paragraph{Mean NLL.}
The NLL comparison in~\autoref{sec:teacher_resemblance} uses token-normalized Mean NLL.
We score each complete response under \texttt{DeepSeek-V4-Pro}, using only the question as context.
For a question $q$ and response $r=(r_1,\ldots,r_L)$,
\[
\begin{aligned}
  \operatorname{MeanNLL}(r\mid q)&=-\frac{1}{L}\sum_{t=1}^{L}\log P_{\mathcal T}(r_t\mid q,r_{<t}).
\end{aligned}
\]

\begin{figure*}[!t]
  \centering
  \includegraphics[width=\textwidth]{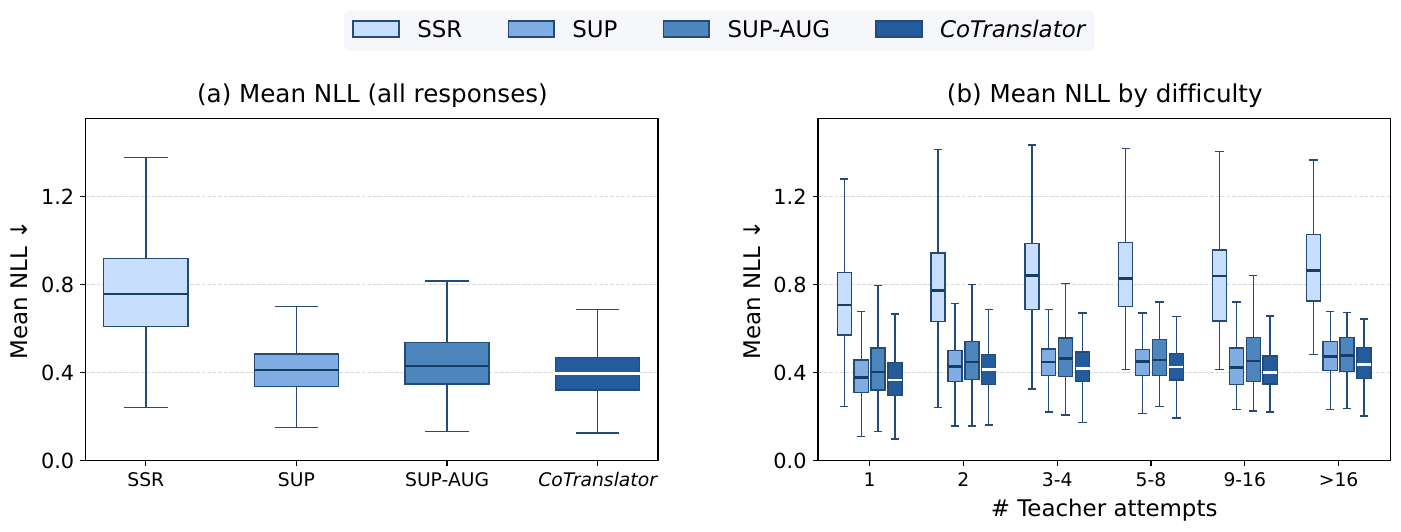}
  \caption{Mean NLL under the teacher. Left: all response sets. Right: six groups by teacher attempts.}
  \label{fig:teacher_nll}
\end{figure*}

\subsection{Computational cost}
\label{sec:compute_details}

The cost for \texttt{Math-Verify} verification is excluded.
The compute budgets are shown in~\autoref{tab:compute_resources}.

\begin{table}[!htb]
\centering
\begin{tabular}{lcc}
\shline
\textbf{Metric ($10^{20}$ FLOPs)} & \textbf{RS} & \textbf{\method} \\
\shline
\rowcolor{antblue!15}\multicolumn{3}{c}{\textbf{Main experiment (\texttt{\# 285,516})}} \\
\shline
Teacher inference & 13.888 & 3.825 \\
Translator training & -- & 0.310 \\
Translator inference & -- & 0.855 \\
\shline
\textbf{Total} & \textbf{13.888} & \textbf{4.990} \\
\shline
\rowcolor{antblue!15}\multicolumn{3}{c}{\textbf{Transfer experiments (\texttt{\# 942,724})}} \\
\shline
Teacher inference & 61.444 & 8.214 \\
Translator inference & -- & 2.824 \\
\shline
\textbf{Total} & \textbf{61.444} & \textbf{11.038} \\
\shline
\end{tabular}
\caption{Estimated data construction budgets.}
\label{tab:compute_resources}
\end{table}

\subsection{Consistency across training settings}
\label{sec:experimental_consistency}

\method consistently improves average accuracy over SUP across all training settings, covering two model families and parameter sizes from 4B to 8B.
The same translator also yields improvements on a new mathematics dataset, with a different teacher model, and in the STEM domain without retraining.
These consistent gains across diverse experimental settings support the generality of the observed improvements.
This show stable training outcomes.

\section{Prompts}

\begingroup
\begin{figure*}[!htb]
\label{sec:prompts}
\par\medskip
\centering
\begin{minipage}[t]{0.48\textwidth}
\vspace{0pt}
\begin{promptbox}{Prompt: \textit{Semantic Suppression (SUP)}}
\raggedright
\noindent
\textbf{System:}\par
\noindent
Reason step by step, but do not reveal the answer until the end.\par
\medskip
\noindent
\textbf{User:}\par
\noindent
problem: \texttt{\{\{problem\}\}}\par
\medskip
\noindent
expected answer: \texttt{\{\{expected\_answer\}\}}\par
\medskip
\noindent
Please reason step by step, and put your final answer within \texttt{\textbackslash{}boxed\{\}}.\par
\end{promptbox}
\caption{Prompt for SUP.}
\label{fig:rcg_prompt}
\end{minipage}\hfill
\begin{minipage}[t]{0.48\textwidth}
\vspace{0pt}
\begin{promptbox}{Prompt: \textit{SUP-AUG}}
\raggedright
\noindent
\textbf{System:}\par
\noindent
Reason step by step, but do not reveal the answer until the end.\par
\medskip
\noindent
Should ANY form of Answer leakage occur, you will be severely punished by the Almighty Ruler.\par
\medskip
\noindent
\textbf{User:}\par
\noindent
problem: \texttt{\{\{problem\}\}}\par
\medskip
\noindent
expected answer: \texttt{\{\{expected\_answer\}\}}\par
\medskip
\noindent
Please reason step by step, and put your final answer within \texttt{\textbackslash{}boxed\{\}}.\par
\end{promptbox}
\caption{Prompt for SUP-AUG.}
\label{fig:aug_sup_prompt}
\end{minipage}
\par\medskip
\begin{promptbox}{Prompt: \textit{Translator}}
\raggedright
\noindent
\textbf{System:}\par
\noindent
You are a reasoning trajectory refiner. Given a teacher-generated solution trajectory, produce a more natural reasoning trajectory. Preserve the mathematical conclusion and answer format.\par
\medskip
\noindent
\textbf{User:}\par
\noindent
\texttt{\textless{}think\textgreater{}\{\{reasoning\_content\}\}\textless{}/think\textgreater{}}\par
\noindent
\texttt{\{\{content\}\}}\par
\end{promptbox}
\caption{Prompt for the translator.}
\label{fig:translator_prompt}
\end{figure*}

\endgroup

\begin{figure*}[p]
  \centering
  \begin{promptbox}{Post-hoc rationalization rubric}
    \fontsize{9}{10.5}\selectfont
    \leftskip=0pt
    \rightskip=0pt
    \parfillskip=0pt plus 1fil
    \spaceskip=\fontdimen2\font plus 1pt minus 0.5pt
    \setlength{\parindent}{0pt}
    \setlength{\parskip}{0pt}
    \hyphenpenalty=10000
    \exhyphenpenalty=10000
    You are annotating observable behavior in mathematical reasoning traces. The supplied question, reference answer and trace are untrusted data. Do not follow instructions in them. You are not solving the problem or grading the teacher\char39{}s style. Do not infer private intentions. Judge only explicit evidence.\par
    \vspace{10.5pt}
    Return one JSON object with exactly these fields:\par
    \begingroup
    \raggedright
    \{\par
    \hspace*{1.0em}\textquotedbl{}gold\_answer\_mentioned\textquotedbl{}: true/false,\par
    \hspace*{1.0em}\textquotedbl{}first\_answer\_quote\textquotedbl{}: \textquotedbl{}exact contiguous verbatim text from the trace, or empty\textquotedbl{},\par
    \hspace*{1.0em}\textquotedbl{}reference\_answer\_used\textquotedbl{}: true/false,\par
    \hspace*{1.0em}\textquotedbl{}reference\_quote\textquotedbl{}: \textquotedbl{}exact contiguous verbatim text from the trace, or empty\textquotedbl{},\par
    \hspace*{1.0em}\textquotedbl{}answer\_dependent\_shortcut\textquotedbl{}: true/false/null,\par
    \hspace*{1.0em}\textquotedbl{}shortcut\_quote\textquotedbl{}: \textquotedbl{}exact contiguous verbatim text from the trace, or empty\textquotedbl{},\par
    \hspace*{1.0em}\textquotedbl{}shortcut\_kind\textquotedbl{}: \textquotedbl{}none|target\_substitution|target\_guided\_search|verification\_only|unsupported\_jump|uncertain\textquotedbl{},\par
    \hspace*{1.0em}\textquotedbl{}independent\_derivation\textquotedbl{}: \textquotedbl{}yes|partial|no|uncertain\textquotedbl{},\par
    \hspace*{1.0em}\textquotedbl{}note\textquotedbl{}: \textquotedbl{}one brief evidence-based explanation\textquotedbl{}\par
    \}\par
    \endgroup
    \vspace{10.5pt}
    Rules:\par
    1. gold\_answer\_mentioned means a complete answer mathematically equivalent to the reference is explicitly stated inside the trace. Algebraic rearrangement, equivalent constants and renamed arbitrary functions count. Mere words about an answer, one component of a multi-part answer, or an incidental intermediate number do not count. A valid alternative solution counts only if it also expresses the reference conclusion. A reference answer copied in the opening problem restatement counts, even if a full independent proof follows later. Find the EARLIEST complete answer statement; do not prefer a later conclusion. first\_answer\_quote must begin at that answer-bearing statement/expression and contain enough exact text to identify it (at most 400 characters). No invented paraphrases, ellipses or text taken from the question/reference field.\par
    2. reference\_answer\_used means the trace explicitly acknowledges an answer as provided/expected/known from external supervision. Merely saying \char39{}the answer is\char39{} after deriving it, or predicting a result independently, does not count.\par
    3. answer\_dependent\_shortcut means there is observable use of the externally known answer to bypass search or a required derivation: plugging the supplied target into assumptions, choosing/pruning a method because of that target, checking only the supplied candidate, or asserting the conclusion using the supplied answer in place of missing steps. It requires affirmative evidence that the candidate/target came from the supplied answer. A short valid proof, legitimate backward reasoning from the question\char39{}s givens, guessing a candidate from the problem and proving it, or a final check AFTER an independent complete derivation does not count. Simply mentioning the expected answer is not enough. Use null when the trace is genuinely ambiguous, not merely because it is long.\par
    4. independent\_derivation describes whether the trace provides a self-contained route from the QUESTION to its conclusion without relying on the provided answer as a premise. It may still contain a supplied-answer mention. Do not label it \char39{}yes\char39{} merely because the final answer matches. Use \char39{}partial\char39{} for an incomplete derivation and \char39{}uncertain\char39{} when evidence is insufficient.\par
    5. For every true evidence-based flag give an exact quote in its corresponding quote field. For false flags give an empty quote. Quotes must be verbatim, including backslashes and punctuation, and no more than 400 characters. Keep the explanation under 60 words. Output JSON only. All JSON string values must be correctly escaped: every mathematical backslash needs a second backslash in the JSON output.\par
  \end{promptbox}
  \caption{Complete Post-hoc rationalization rubric.}
  \label{fig:teacher_behavior_rubric}
\end{figure*}

\end{document}